\documentclass[10pt,twocolumn,letterpaper]{article}

\usepackage[pagenumbers]{cvpr} 

\definecolor{cvprblue}{rgb}{0.21,0.49,0.74}
\usepackage[pagebackref,breaklinks,colorlinks,allcolors=cvprblue]{hyperref}

\usepackage{booktabs}
\usepackage{multirow}
\usepackage[table]{xcolor}
\usepackage{colortbl}
\usepackage{pifont} 
\usepackage{makecell}
\usepackage[most]{tcolorbox}

\newcommand{\cmark}{\ding{51}} 
\newcommand{\xmark}{\ding{55}} 

\def\paperID{18009} 
\def\confName{CVPR}
\def\confYear{2026}

\title{AgentComp: From Agentic Reasoning to Compositional Mastery in Text-to-Image Models}

\author{Arman Zarei\textsuperscript{1,2,*},\hspace{2pt} Jiacheng Pan\textsuperscript{1},\hspace{2pt} Matthew Gwilliam\textsuperscript{1},\hspace{2pt} Soheil Feizi\textsuperscript{2},\hspace{2pt} Zhenheng Yang\textsuperscript{1}\\
  \textsuperscript{1}TikTok \hspace{14pt}
  \textsuperscript{2}University of Maryland
}

\begin{document}
\twocolumn[{%
    \renewcommand\twocolumn[2][]{#1}
    \maketitle
    \iftoggle{cvprfinal}{\vspace{-1cm}}{\vspace{-1.2cm}}
    \begin{center}
        \centering \centering
        \includegraphics[width=\iftoggle{cvprfinal}{0.90}{0.95}\textwidth]{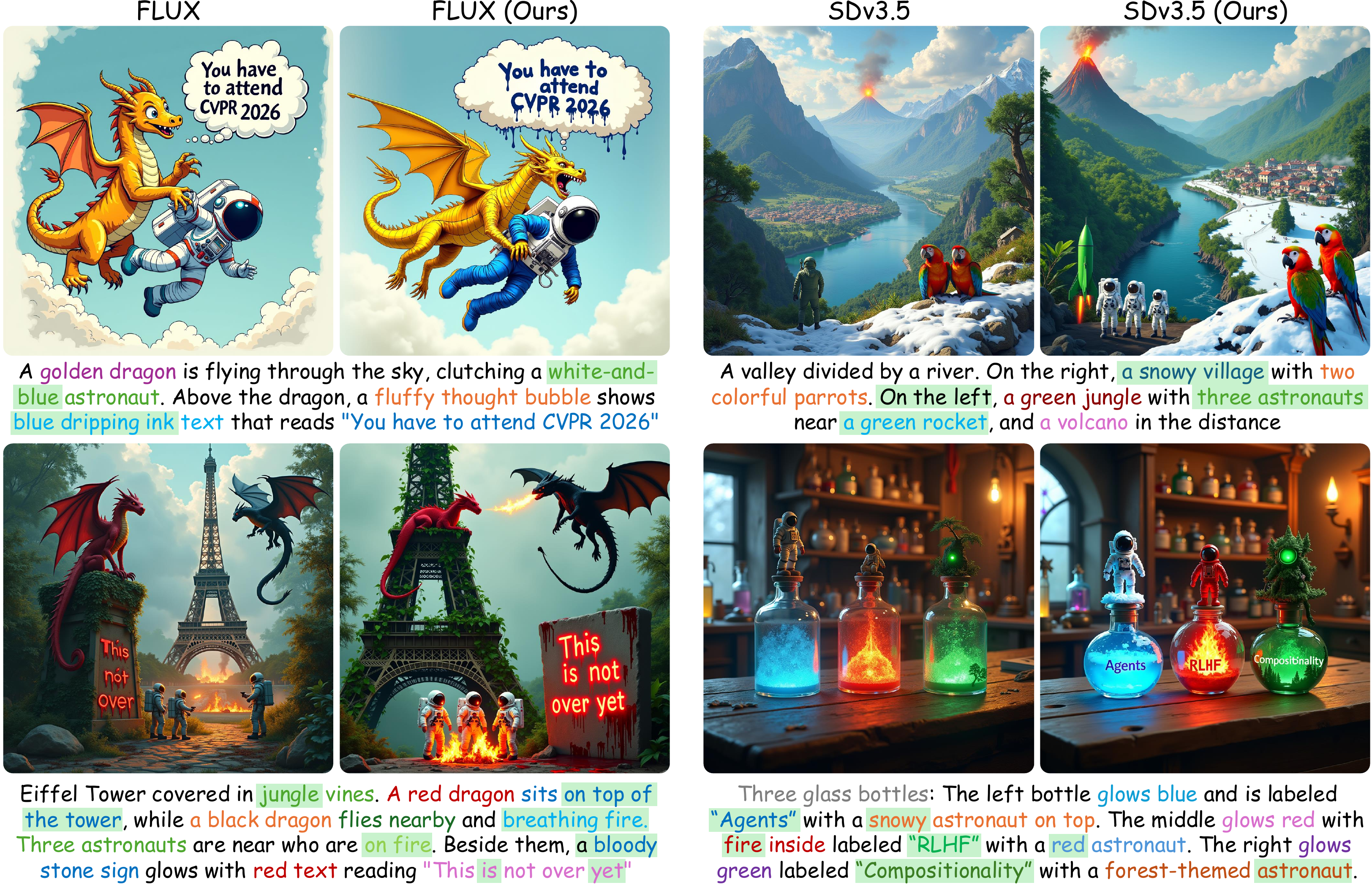}
        \vspace{-0.15cm}
        \captionof{figure}{AgentComp significantly enhances the compositional abilities of text-to-image generative models, improving text–image alignment while preserving image quality and even boosting capabilities such as text rendering, despite not being explicitly trained for it.}
    \label{fig:teaser}
    \end{center}
}] 

\begin{abstract}
Text-to-image generative models have achieved remarkable visual quality but still struggle with compositionality—accurately capturing object relationships, attribute bindings, and fine-grained details in prompts. 
A key limitation is that models are not explicitly trained to differentiate between compositionally similar prompts and images, resulting in outputs that are close to the intended description yet deviate in fine-grained details.
To address this, we propose AgentComp, a framework that explicitly trains models to better differentiate such compositional variations and enhance their reasoning ability. AgentComp leverages the reasoning and tool-use capabilities of large language models equipped with image generation, editing, and VQA tools to autonomously construct compositional datasets. Using these datasets, we apply an agentic preference optimization method to fine-tune text-to-image models, enabling them to better distinguish between compositionally similar samples and resulting in overall stronger compositional generation ability.
AgentComp achieves state-of-the-art results on compositionality benchmarks such as T2I-CompBench, without compromising image quality—a common drawback in prior approaches—and even generalizes to other capabilities not explicitly trained for, such as text rendering. 
\iftoggle{cvprfinal}{
\begingroup
\renewcommand\thefootnote{\fnsymbol{footnote}}
\footnotetext[1]{Work done during internship at TikTok}
\endgroup
}{}
\iftoggle{cvprfinal}{
we will open-source the checkpoints, data, and codebase to support future research.\footnote[1]{Project page is available at: \href{https://armanzarei.github.io/AgentComp}{https://armanzarei.github.io/AgentComp}}
}{}
\end{abstract}

\section{Introduction}
\label{sec:intro}

Recent advancements in diffusion~\cite{ho2020denoising, nichol2021improved, song2020score, rombach2022high} and flow-matching models~\cite{lipman2022flow, esser2024scaling} have led to impressive progress in text-to-image generation, as seen in models such as FLUX~\cite{flux2024, labs2025flux}, Stable Diffusion~\cite{esser2024scaling, podell2023sdxl}, Qwen-Image~\cite{wu2025qwen}, and DALL-E \cite{ramesh2021zero}. These models can synthesize highly realistic and text-aligned images. However, they still struggle to capture fine-grained compositional details~\cite{huang2023t2i, feng2022training, liu2022compositional}—for example, binding attributes to the correct objects or preserving spatial and relational consistency among multiple entities within a scene.

Several recent works have attempted to mitigate this limitation. Some optimize the attention mechanisms within diffusion models to better capture the semantics of the text prompt \cite{chefer2023attend, meral2024conform, rassin2023linguistic, bao2024separate, jiang2024comat}, while others use LLMs as planners to guide the generation process \cite{li2023gligen, yang2024mastering, zhang2024realcompo, liu2024draw, li2025mccd}, or fine-tune the model on high-quality compositional datasets \cite{huang2023t2i, zhang2024itercomp, zarei2024understanding, li2024all, zarei2024improving}. Although these approaches improve over the baselines, they often fail to address the underlying cause of compositional errors, exhibit limited generalization to unseen distributions, and tend to degrade image quality or other core capabilities of the model.

In this paper, we take a step back and hypothesize that the core reason behind poor compositional performance lies in the lack of explicit training signal to distinguish between samples that are visually similar but differ in subtle compositional details (Fig.~\ref{fig:motivation}). This results in outputs that are often close to the intended description yet deviate in fine-grained compositional details. We argue that if the model is explicitly guided to differentiate such closely related compositional trajectories, its ability to generate compositionally correct images can improve substantially.

To this end, we introduce AgentComp, an agentic framework that autonomously constructs a high-quality compositional dataset by generating groups of near-identical yet compositionally contrasted samples and leverages it to fine-tune text-to-image models. Manually curating such datasets is labor-intensive; therefore, AgentComp employs agentic LLMs equipped with reasoning and tool-use capabilities to automatically create compositional prompt–image pairs. Through iterative reasoning and tool calls, the agent generates fine-grained contrastive samples that differ only in small compositional details. Using this dataset, AgentComp applies Agentic Preference Optimization to refine the model’s denoising trajectories, explicitly encouraging the model to prefer more compositionally faithful generations.

Through extensive quantitative and qualitative evaluations, we show that AgentComp significantly enhances the compositional ability of base models—achieving state-of-the-art performance on T2I-CompBench when built upon FLUX. Moreover, unlike many prior fine-tuning methods that degrade image quality or other capabilities, AgentComp preserves and in some cases improves general image quality and even enhances text-rendering capabilities.

In summary, our main contributions are:
\begin{itemize}
    \item We propose a fully autonomous agentic framework that constructs high-quality compositional datasets consisting of visually similar images differing only in subtle details, used to improve compositional reasoning in T2I models.
    \item We introduce an agentic preference optimization method that explicitly trains text-to-image models to follow compositionally correct trajectories and to avoid visually similar but compositionally incorrect alternatives.
    \item AgentComp achieves state-of-the-art results on T2I-CompBench while maintaining overall quality. It also generalizes well and substantially improves capabilities like text rendering, even without relevant training data.
\end{itemize}

\section{Related Works}
\label{sec:related_works}


\subsection{Compositionality in Text-to-Image Models}

Previous efforts to enhance compositionality in text-to-image models have explored multiple directions and made notable progress, yet important limitations remain. Some methods \cite{chefer2023attend, meral2024conform, rassin2023linguistic, bao2024separate, jiang2024comat} directly intervene in the attention mechanism by modifying or optimizing attention maps. However, they typically target narrow aspects of compositionality—such as attribute binding—and do not generalize well to others, like spatial relationships. Planning-based approaches \cite{li2023gligen, yang2024mastering, zhang2024realcompo, liu2024draw, li2025mccd} instead introduce an explicit scene-structuring stage, where an intermediate layout or structural plan is generated (often via an LLM or user input) to guide the subsequent image synthesis process. Although effective in certain scenarios, they add noticeable computational overhead at inference-time and only offer a temporary workaround rather than directly addressing the model’s inherent compositional limitations. Other methods focus on fine-tuning with curated datasets \cite{huang2023t2i, zhang2024itercomp, zarei2024understanding, li2024all, han2024progressive, zarei2024improving}, enabling models to learn more accurate attribute and relationship bindings. However, these can overly alter the base model’s behavior, leading to degraded image quality \cite{zarei2024improving, han2024progressive} or loss of capabilities such as text rendering. 
In contrast, our work leverages an agentic orchestrator that performs reasoning-driven synthesis of groups of near-identical images that vary only in subtle compositional details, and applies an agentic preference optimization objective that steers the model away from compositionally incorrect yet visually similar trajectories, yielding substantial compositional improvements while preserving overall quality and even enhancing capabilities such as text rendering.

\subsection{Agentic AI and Tool Use}


Agentic AI frames an LLM as a planner–executor that decomposes tasks into tool calls and iteratively verifies outcomes, enabling reliable, multi-step problem solving. Early formulations \cite{yao2023react} unify reasoning and acting to ground tool use in chain-of-thought \cite{wei2022chain}, while subsequent systems \cite{schick2023toolformer, patil2024gorilla, shen2023hugginggpt} learn or orchestrate tool APIs at scale to expand an agent’s operational repertoire. In the multimodal setting, agents coordinate specialist vision models—planning and invoking segmentation, detection, or captioning tools—and even external code, to solve complex visual tasks \cite{yang2023mm, wu2023visual, hu2024visual}. Parallel lines use agentic pipelines for automatic dataset construction and instruction expansion \cite{mitra2024agentinstructgenerativeteachingagentic}. 
Our work leverages the reasoning and tool-use capabilities of agentic orchestration to autonomously mine highly informative, contrastive, and compositionally controlled samples through coordinated use of specialized generative and evaluative tools, and then uses these samples to explicitly enhance compositional reasoning in T2I models.
\section{Method}
\label{sec:method}

In this section, we first present the motivation behind our approach (Sec.~\ref{sec:motivation}) and the need for an agentic framework for generating compositional data.
We then describe the proposed agentic data generation pipeline (Sec.~\ref{sec:agentic_dataset_generation}), where coordinated agents collaboratively create contrastive samples.
Finally, we explain how the resulting data are used in our preference optimization stage to enhance compositionality in text-to-image models (Sec.~\ref{sec:apo}).
Together, these components form a unified framework that achieves state-of-the-art performance on compositional benchmarks.

\subsection{Motivation}
\label{sec:motivation}

Recent text-to-image diffusion and flow models have achieved remarkable generative quality by being trained on massive text–image corpora. However, these trainings lack explicit supervision for distinguishing between images that are visually similar yet compositionally different. Consequently, these models often generate visually coherent yet compositionally incorrect images, failing to capture fine-grained relationships such as object–attribute bindings, spatial layouts, and counting details.

\begin{figure}[t]
    \centering
    \includegraphics[width=\linewidth]{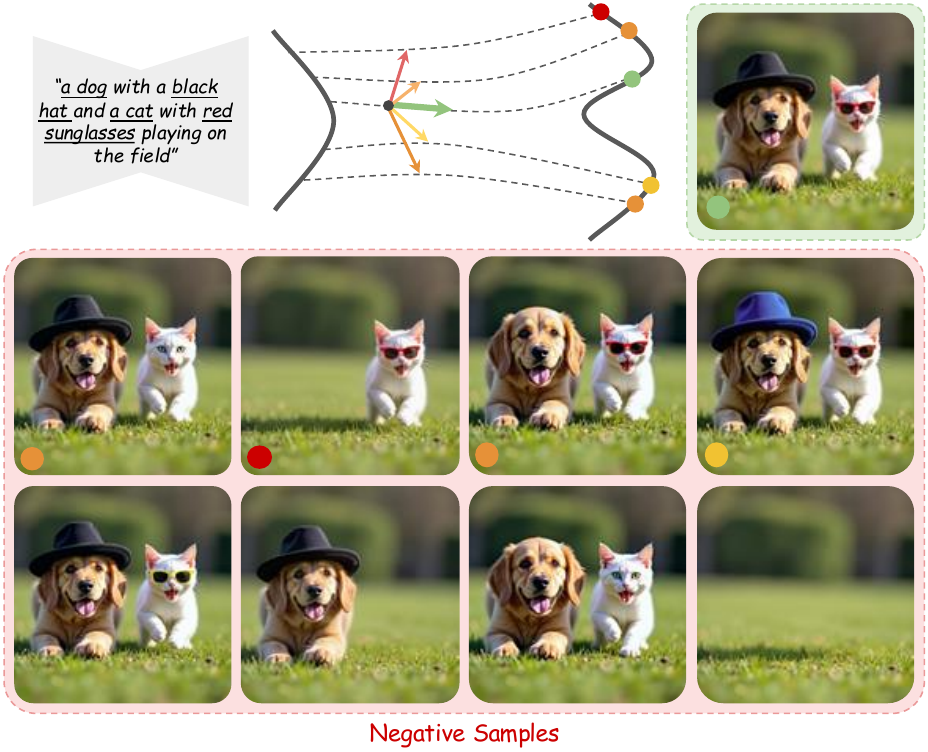}
    \vspace{-0.7cm}
    \caption{\textbf{Motivation for correcting compositional trajectories.} During the denoising trajectory for a compositional prompt, the model is not explicitly trained to avoid visually similar paths that miss certain compositional details.}
    \label{fig:motivation}
\end{figure}

As illustrated in Fig.~\ref{fig:motivation}, when generating an image for a compositional prompt $\mathcal{P}$, the model’s denoising trajectory may drift toward an alternative path that produces semantically close but compositionally incorrect results. For instance, given the prompt “a dog with a black hat and a cat with red sunglasses playing on a field”, the model may omit some details and generate “a dog with a hat and a cat” instead. This behavior stems from the absence of training signals that explicitly discourage such incorrect trajectories. To address this, we aim to teach the model to recognize and avoid these deviations by constructing a contrastive dataset composed of positive compositional pairs $(\mathcal{P}, \mathcal{I})$ and corresponding negatives $\{(\mathcal{P}_k^-, \mathcal{I}_k^-)\}$—samples that are semantically close to the reference but diverge in specific compositional details. Leveraging these examples, we guide the model to prefer generation trajectories that faithfully follow the intended compositional details of the prompt.

\subsection{Agentic Contrastive Dataset Generation}
\label{sec:agentic_dataset_generation}

Manually building the dataset described above is prohibitively labor-intensive. For a single data point of the dataset, one must (i) generate a compositionally correct image $\mathcal{I}$ for a prompt $\mathcal{P}$ and verify that every detail is satisfied, and (ii) craft negative images $\mathcal{I}^-_k$ paired with negative prompts $\mathcal{P}^-_k$ that are deliberately misaligned with $\mathcal{P}$ (e.g., missing attributes, swapped colors, incorrect counts, or violated spatial relations). Furthermore, scaling this process to thousands of examples is impractical.

We address this by leveraging recent agentic systems that can plan, call tools, and reason over intermediate feedback. Using these agents, we fully automate the construction of the contrastive dataset needed to steer generation trajectories. In the following section, we describe the individual agents that compose our agentic orchestra, and then explain how they collaboratively operate to achieve our goal.

\begin{figure}[t]
    \centering
    \includegraphics[width=\linewidth]{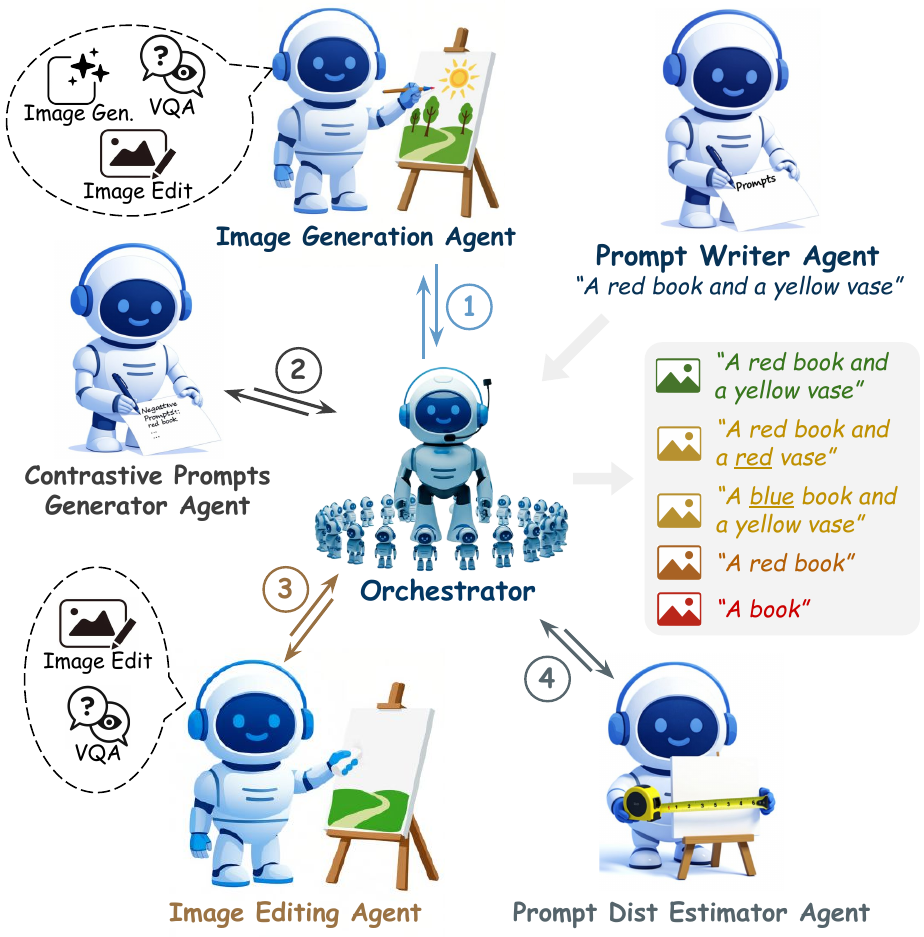}
    \vspace{-0.6cm}
    \caption{\textbf{Illustration of the agentic orchestration.} The orchestrator collaborates with specialized agents to generate a positive image, synthesize contrastive prompts, produce corresponding negative images, and rank them by compositional distance.}
    \label{fig:agent_orchestra}
\end{figure}

\textbf{Agents Orchestra.}
Given an input prompt $\mathcal{P}$, the goal is to produce a corresponding compositionally correct image $\mathcal{I}$, along with a set of negative samples $\{(\mathcal{P}_k^-, \mathcal{I}_k^-)\}$ ranked by their compositional distance from the original pair $(\mathcal{P}, \mathcal{I})$. Figure~\ref{fig:agent_orchestra} provides an overview of the orchestration process. The pipeline begins with a textual prompt $\mathcal{P}$, which can either be manually specified or automatically generated by a language model or the prompt-writer agent. The orchestrator (or coordinator) then manages communication between the agents to accomplish the overall goal.

\begin{figure}[t]
    \centering
    \includegraphics[width=\linewidth]{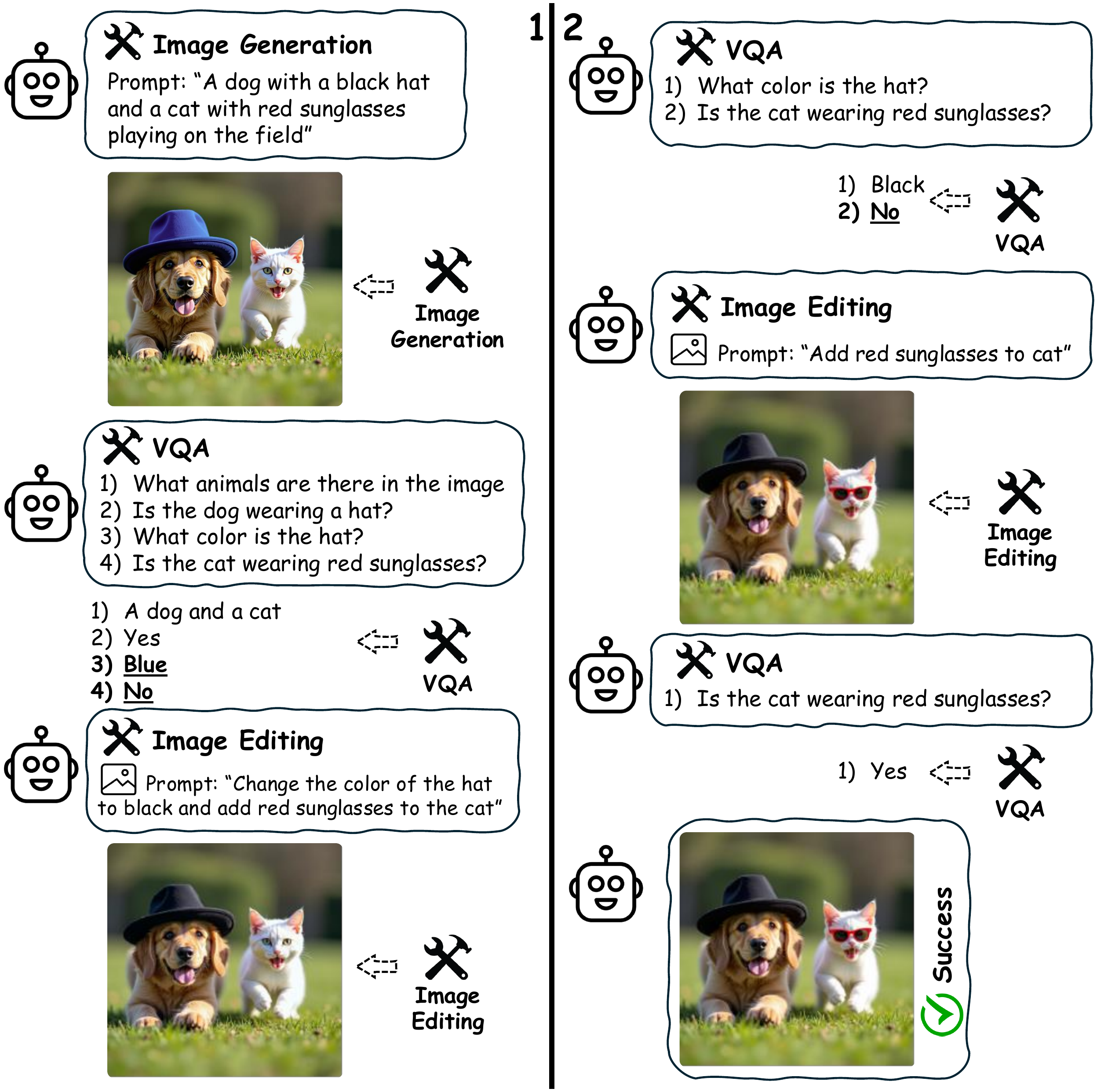}
    \vspace{-0.6cm}
    \caption{\textbf{Example scenario of the Image Generation Agent.} The agent employs iterative reasoning and tool calls to produce a compositionally accurate image that aligns with the given prompt.}
    \label{fig:image_gen_agent_conversation}
\end{figure}

\begin{enumerate}
    \item \emph{Positive Image Generation.} 
    The orchestrator first sends $\mathcal{P}$ to the Image Generation Agent $\mathcal{A}_\text{ImageGen}$, which produces a compositionally accurate image $\mathcal{I}$ aligned with the given prompt.
    This agent is equipped with multiple tools that enable it to iteratively refine the generated image until all compositional details are satisfied. Specifically, it has access to:
    \begin{itemize}
        \item an \emph{image generation model}, which creates an initial image based on $\mathcal{P}$, but may overlook fine-grained attributes or object relations;    
        \item a \emph{visual question answering (VQA)} module, which allows the agent to inspect the generated image, identify missing or incorrect elements, and reason about how it deviates from the prompt; and
        \item an \emph{image editing model}, capable of applying targeted edits to correct these discrepancies.
    \end{itemize}
    Through iterative coordination between the VQA and editing tools, the agent continuously verifies and adjusts the image—re-generating or refining local regions when necessary—until the final output $\mathcal{I}$ satisfies the compositional constraints of the prompt with high fidelity. Figure~\ref{fig:image_gen_agent_conversation} illustrates a sample scenario in which $\mathcal{A}_\text{ImageGen}$ employs multiple rounds of reasoning and tool calls to generate an image for a given compositional prompt.
    \item \emph{Negative Prompt Construction.}
    In parallel, the Contrastive Prompt Generation Agent $\mathcal{A}_\text{ContraPromptGen}$ synthesizes a set of contrastive prompts ${\mathcal{P}^-_k}$ by introducing controlled perturbations—e.g., removing attributes, swapping relations, or altering object properties—to form meaningful compositional negative samples.
    \item \emph{Negative Image Generation.}
    Each contrastive prompt $\mathcal{P}_k^-$, along with the reference image $\mathcal{I}$, is then passed to the Image Editing Agent $\mathcal{A}_\text{ImageEdit}$, which generates a corresponding image $\mathcal{I}_k^-$ such that the perturbation from $\mathcal{I}\rightarrow\mathcal{I}^-_k$ aligns with the semantic change from $\mathcal{P}\rightarrow\mathcal{P}^-_k$. Similar to $\mathcal{A}_\text{ImageGen}$, $\mathcal{A}_\text{ImageEdit}$ is equipped with the necessary tools—such as VQA and image editing tools—to accurately analyze and execute the required modifications.
    \item \emph{Compositional Distance Estimation.}
    Once all pairs $(\mathcal{P}_k^-,\mathcal{I}_k^-)$ are obtained, they are evaluated by the Distance Estimator Agent $\mathcal{A}_\text{DistEst}$, which assigns a compositional distance score $d_k$ to each negative sample, measuring how far it deviates from the original pair $(\mathcal{P},\mathcal{I})$ by estimating the number of atomic modifications required to transform one into the other. Here, atomic refers to minimal compositional edits—such as changing a single attribute, incrementally adjusting object count, or swapping a localized spatial relation.
\end{enumerate}
The orchestrator then compiles the final output, consisting of the positive pair and its ranked set of negatives:

\[
(\mathcal{P}, \mathcal{I}), \{(\mathcal{P}_k^-, \mathcal{I}_k^-, d_k)\}_{k=1}^K.
\]

\begin{figure}[t]
    \centering
    \includegraphics[width=\linewidth]{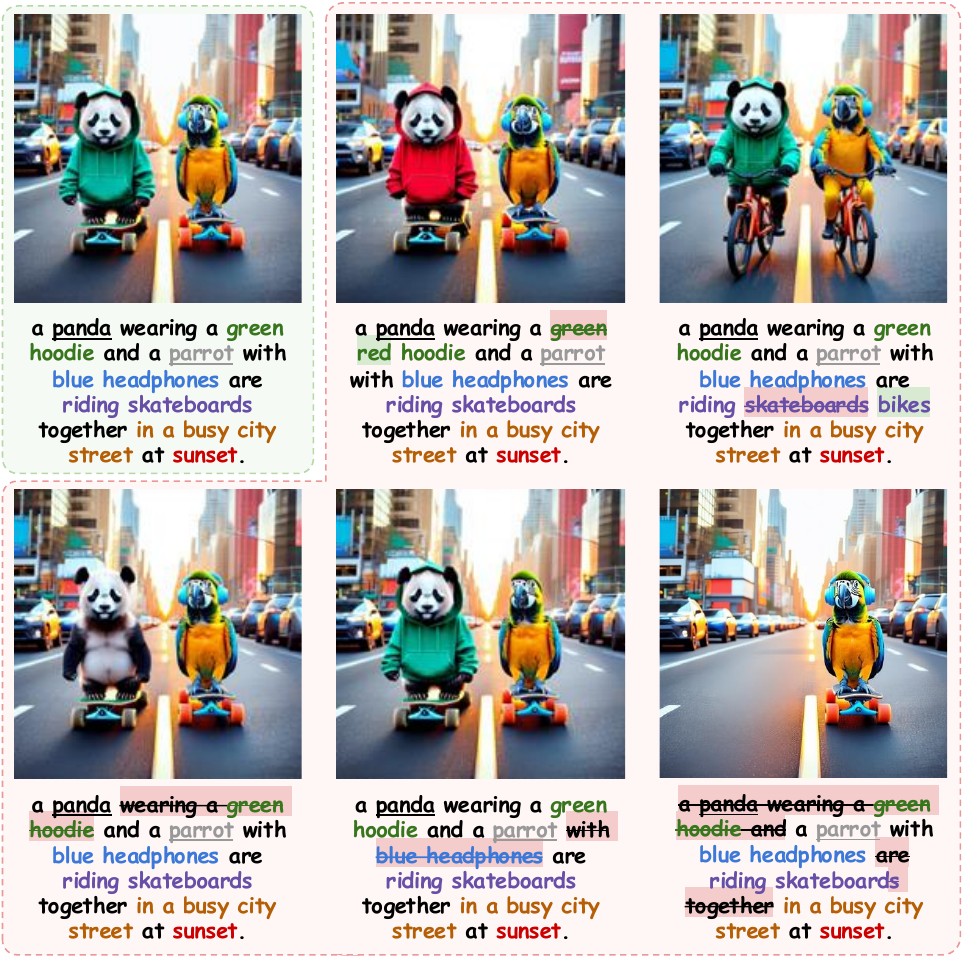}
    \vspace{-0.6cm}
    \caption{\textbf{Example from the dataset generated by the agentic orchestra.} The dataset includes high-quality samples, with reference image that accurately capture compositional details in the given prompt, along with negative samples created by subtly altering those details in the reference text–image pair.}
    \label{fig:dataset_generation_sample}
\end{figure}

Figure~\ref{fig:dataset_generation_sample} illustrates an example of the dataset generated by the agentic orchestra. Repeating this procedure across diverse compositional prompts yields a dataset that provides the foundation for training models to avoid incorrect trajectories and promote compositionally faithful outputs.

\subsection{Agent Preference Optimization}
\label{sec:apo}

Having the aforementioned dataset, we now apply preference optimization to explicitly guide the model toward compositionally accurate generations. Following the standard formulation in the preference learning literature, each sample in our dataset contains a ranked pair under a given conditioning prompt: $\mathcal{I}^+ \succ \mathcal{I}^- | \mathcal{P}$ (for some $\mathcal{I}^-$ in $\{\mathcal{I}^-_k\}_{k=1}^K$). According to the Bradley–Terry model \cite{Bradley1952RANKAO}, the probability of preferring $\mathcal{I}^+$ over $\mathcal{I}^-$ is defined as:
$$
p_\text{BT}(\mathcal{I}^+ \succ \mathcal{I}^- | \mathcal{P}) = \sigma (r(\mathcal{P}, \mathcal{I}^+) - r(\mathcal{P}, \mathcal{I^-})).
$$
where $r(\mathcal{P}, \mathcal{I})$ represents a reward function parameterized by a neural network $\psi$. The reward model can be learned via maximum-likelihood estimation:
$$
\mathcal{L}_\text{BT}(\psi) = - \mathbb{E}_{\mathcal{P},\mathcal{I}^+, \mathcal{I}^-}[\log \sigma (r_\psi(\mathcal{P}, \mathcal{I}^+) - r_\psi(\mathcal{P}, \mathcal{I}^-))]
$$
In reinforcement learning from human feedback (RLHF), the goal is to learn a conditional distribution $p_\theta(\mathcal{I}|\mathcal{P})$ that maximizes the expected reward while remaining close to a reference model $p_\text{ref}(\mathcal{I}|\mathcal{P})$. This can be expressed as:
\begin{align*}
\max_{p_\theta} \;\; & \mathbb{E}_{\mathcal{P}\sim \mathcal{D}_\mathcal{P}, \mathcal{I}\sim p_\theta(\mathcal{I}|\mathcal{P})}[r(\mathcal{P}, \mathcal{I})] \\
&- \beta \mathbb{D}_\text{KL}[p_\theta(\mathcal{I}|\mathcal{P}) \| p_\text{ref}(\mathcal{I}|\mathcal{P})]
\end{align*}

Recent works \cite{rafailov2023direct, wallace2024diffusion} show that, instead of explicitly training a reward model and performing RL, one can directly optimize the model parameters through a reparameterized objective. In diffusion models, this leads to the loss:
\begin{equation*}
\small
\begin{aligned}
\mathcal{L}(\theta) &= - \mathbb{E}_{(\mathcal{I}^+, \mathcal{I}^-, \mathcal{P}) \sim \mathcal{D}} \Big[ 
 \log \sigma \big(-\beta T \,  \, ( \ell^+ - \ell^- ) \big)  \Big] \\
\ell^+ &= \| \epsilon^+ - \epsilon_\theta(\mathcal{I}^+_t, \mathcal{P}, t) \|_2^2 - \| \epsilon^+ - \epsilon_{\text{ref}}(\mathcal{I}^+_t, \mathcal{P}, t) \|_2^2 \\
 \ell^- &= \| \epsilon^- - \epsilon_\theta(\mathcal{I}^-_t, \mathcal{P}, t) \|_2^2 - \| \epsilon^- - \epsilon_{\text{ref}}(\mathcal{I}^-_t, \mathcal{P}, t) \|_2^2 
\end{aligned}
\end{equation*}
where $\mathcal{I}^{+/-}_t$ denotes the noisy latent sample at diffusion timestep $t$, $\epsilon_\theta$ represents the trainable denoising network, $\epsilon_\text{ref}$ the reference (base) model, $\epsilon^{+/-}$ the ground-truth denoising target, and the expectation is taken over $t\sim\mathcal{U}[0,1]$.
For a full derivation, please refer to Appendix~\ref{sec:app:apo}.

This formulation treats all negative samples equally within a ranked group $\{\mathcal{I}^-_k\}_{k=1}^K$, regardless of their compositional proximity to the positive sample.
To better leverage the fine-grained distances estimated by our agentic orchestration, we reweight the loss by introducing a distance-dependent scaling function $\mathcal{H}(.)$ applied to $\beta$. By reweighting $\beta$, we intuitively tune how freely the model can deviate from the reference to satisfy the preference.
Our final Agent Preference Optimization (APO) objective becomes:
\vspace{5pt}
\begin{equation*}
\footnotesize
\begin{aligned}
\mathcal{L}_\text{APO}(\theta) = - \mathbb{E}_{(\mathcal{I}^+, \mathcal{I}_k^-, \mathcal{P}, d_k) \sim \mathcal{D}} \Big[ 
 \log \sigma \big(- \mathcal{H}(d_k) \beta T \,  \, ( \ell^+ - \ell^- ) \big)  \Big]
\end{aligned}
\vspace{5pt}
\end{equation*}
This distance-aware weighting offers finer control over sample importance, resulting in faster and more reliable convergence during fine-tuning.

\section{Experiments and Results}
\label{sec:experiments}

We conduct comprehensive quantitative and qualitative evaluations of AgentComp, demonstrating its robust performance across a wide range of metrics while preserving the original model’s visual quality. Our results show that AgentComp achieves state-of-the-art performance on compositionality benchmarks and exhibits strong generalization to unseen tasks, delivering overall superior results across diverse evaluation settings.

\begin{table*}[h]
\centering
\small
\caption{Quantitative comparison of AgentComp against other baselines on T2I-CompBench.}
\vspace{-0.2cm}
\renewcommand{\arraystretch}{0.9}
\setlength{\tabcolsep}{4pt}
\begin{tabular}{lccccccccc}
\toprule
\multirow{2}{*}{\textbf{Model}} & \multicolumn{3}{c}{\textbf{Attribute Binding}} & \multicolumn{2}{c}{\textbf{Object Relationship}} & \multirow{2}{*}{\textbf{Numeracy $\uparrow$}} & \multirow{2}{*}{\textbf{Complex $\uparrow$}} \\
\cmidrule(lr){2-4} \cmidrule(lr){5-6}
 & \textbf{Color $\uparrow$} & \textbf{Shape $\uparrow$} & \textbf{Texture $\uparrow$} & \textbf{Spatial $\uparrow$} & \textbf{Non-Spatial $\uparrow$} & & \\
\midrule
Composable \cite{liu2022compositional} & 0.4063 & 0.3299 & 0.3645 & 0.0800 & 0.2980 & 0.4272 & 0.2898 \\
Structured \cite{feng2022training} & 0.4990 & 0.4218 & 0.4900 & 0.1386 & 0.3111 & 0.4557 & 0.3355 \\
Attn-Excite \cite{chefer2023attend} & 0.6400 & 0.4517 & 0.5963 & 0.1455 & 0.3109 & 0.4773 & 0.3401 \\
GORS \cite{huang2023t2i} & 0.6603 & 0.4785 & 0.6287 & 0.1815 & 0.3193 & 0.4830 & 0.3328 \\
DALL-E 2 \cite{ramesh2021zero} & 0.5750 & 0.5464 & 0.6374 & 0.1283 & 0.3043 & - & 0.3696 \\
SDXL \cite{podell2023sdxl} & 0.6369 & 0.5408 & 0.5637 & 0.2032 & 0.3110 & 0.5078 & 0.4091 \\
PixArt-$\alpha$ \cite{chen2023pixart} & 0.6886 & 0.5582 & 0.7044 & 0.2082 & 0.3179 & - & 0.4117 \\
ConPreDiff \cite{yang2023improving} & 0.7019 & 0.5637 & 0.7021 & 0.2362 & 0.3195 & - & 0.4184 \\
EvoGen \cite{han2024progressive} & 0.7104 & 0.5457 & 0.7234 & 0.2176 & \textbf{0.3308} & - & \underline{0.4252} \\
RPG \cite{yang2024mastering} & 0.6406 & 0.4903 & 0.5597 & 0.2714 & 0.3047 & 0.4742 & 0.3128 \\
T2I-R1 \cite{jiang2025t2i} & \underline{0.8130} & 0.5852 & 0.7243 & 0.3378 & 0.3090 & - & 0.3993 \\
MCCD \cite{li2025mccd} & 0.6278 & 0.4832 &  0.5647 & 0.2350 & 0.3132 & - & 0.3348 \\
FLUX \cite{flux2024} & 0.7736 & 0.5112 & 0.6325 & 0.2747 & 0.3077 & 0.6162 & 0.3622 \\
Qwen-Image \cite{wu2025qwen} & 0.7835 & 0.5401 & 0.6816 & \underline{0.3647} & 0.3109 & \underline{0.6686} & 0.3530 \\
SDv3.5 \cite{esser2024scaling} & 0.7717 & \underline{0.6050} & \underline{0.7250} & 0.2886 & 0.3176 & 0.6327 & 0.3729 \\
\midrule
AgentComp \small{(Ours)} & \textbf{0.8743} & \textbf{0.6681} &  \textbf{0.8142} & \textbf{0.4748} & \underline{0.3196}\footnotemark[1] & \textbf{0.7378} & \textbf{0.4261} \\
\bottomrule
\end{tabular}
\label{tab:model_comparison}

\end{table*}

\begin{figure}[t]
    \vspace{-0.4cm}
    \centering
    \includegraphics[width=\linewidth]{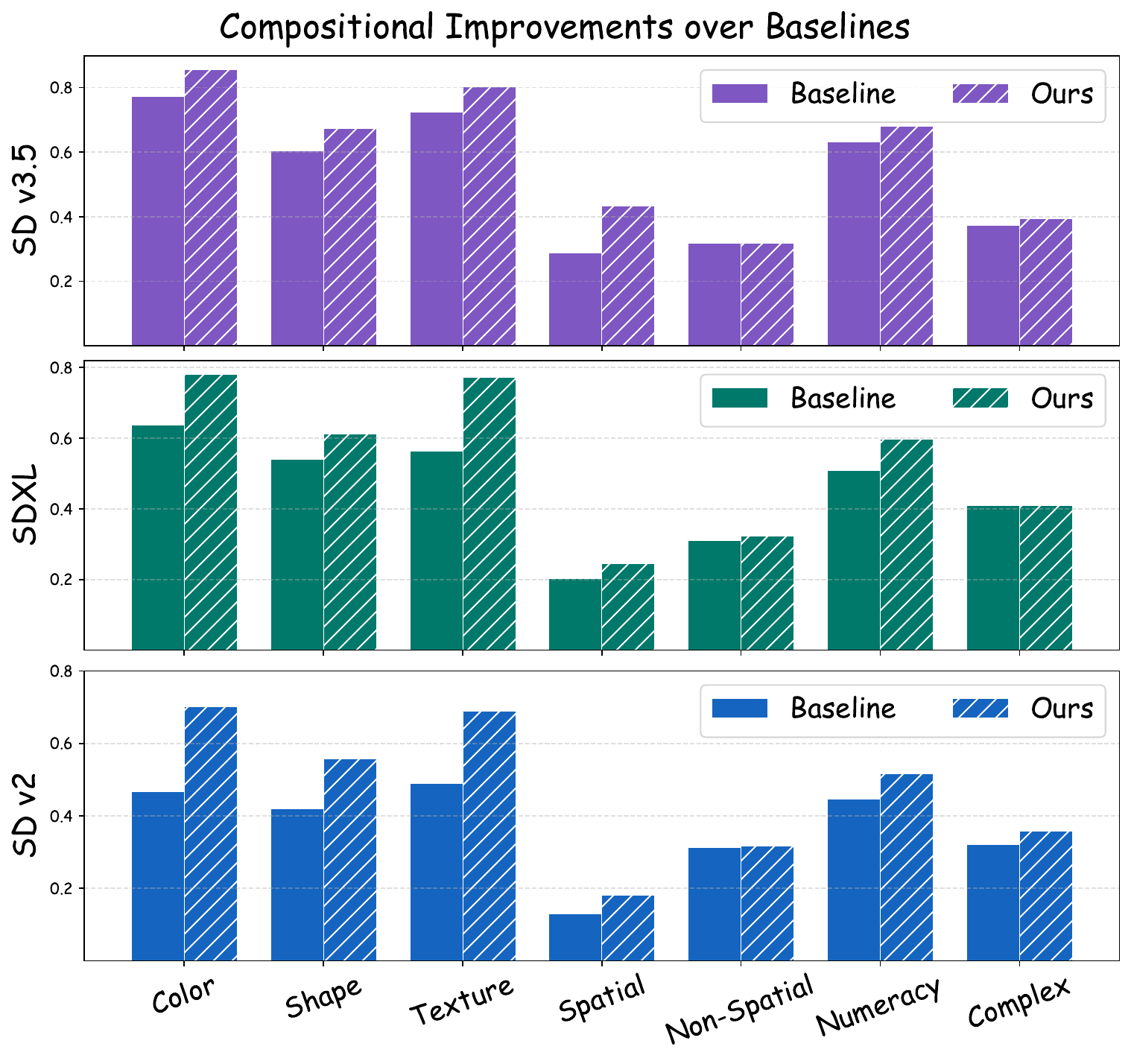}
    \vspace{-0.7cm}
    \caption{\textbf{Performance of AgentComp across base models of varying capacity.} AgentComp shows consistent compositional improvements across all base models, highlighting its robustness and generality.}
    \label{fig:compositional_baselines_improvements}
\end{figure}

\subsection{Implementation Details}
We use FLUX-dev as our main base model in all experiments. To assess the generality and robustness of our approach, we additionally evaluate it on SDv3.5, SDXL, and SDv2, covering models of varying capacities. All models are fine-tuned using LoRA with a rank of 32 for efficiency. For FLUX, we set the base value of $\beta = 100$ in the APO objective and incorporate a normalized distance function within the range [0.5, 1], which adaptively scales $\beta$ between 50 and 100 according to the sample distance. Training is performed at a fixed resolution of $1024\times1024$ with a global batch size of 128 (64 pairs), using mixed-precision (bfloat16) and gradient checkpointing to reduce memory usage and support larger batches. All experiments are conducted on 8×H100 GPUs using distributed training. Further details are provided in Appendix \ref{sec:app:implementation_details}.

\footnotetext[1]{The non-spatial category based on CLIP fails to capture compositional differences, with simple and recent models performing almost identically.}

\begin{figure}[t]
    \vspace{-0.4cm}
    \centering
    \includegraphics[width=\linewidth]{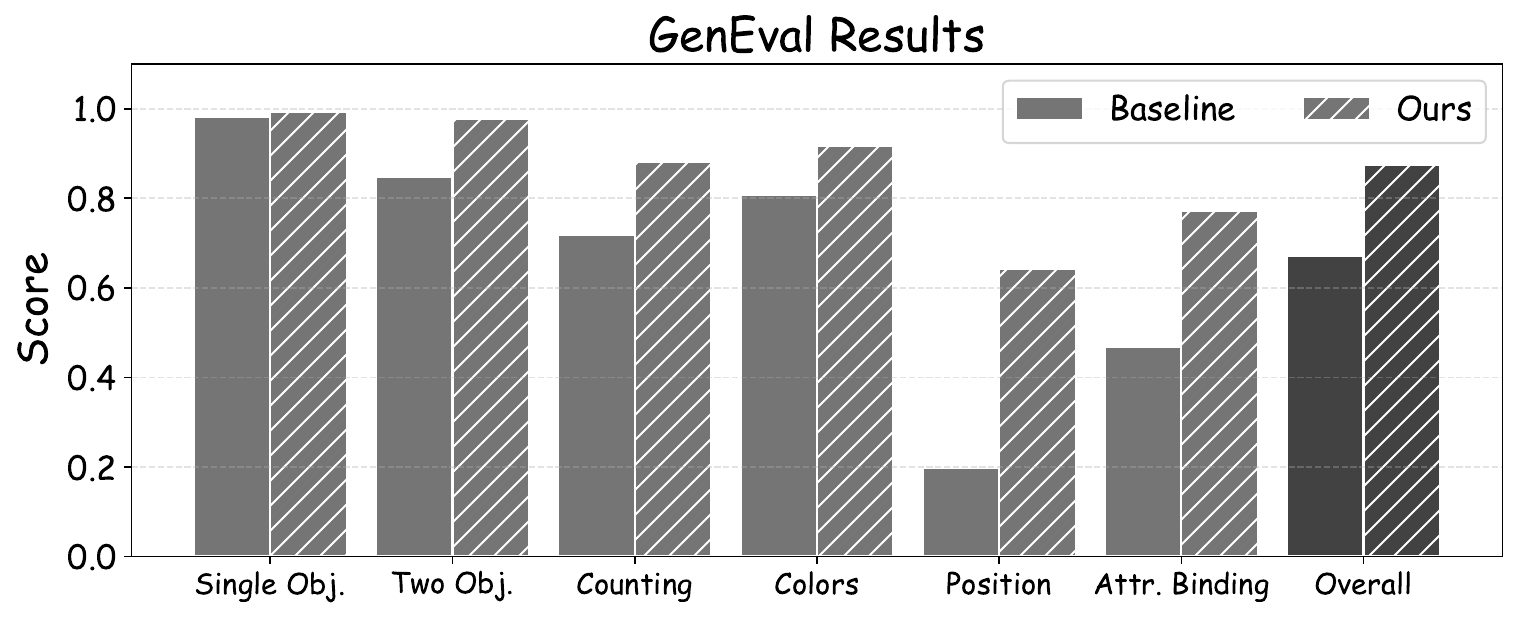}
    \vspace{-0.6cm}
    \caption{\textbf{Performance of AgentComp on GenEval.} AgentComp consistently outperforms the baseline across all categories.}
    \label{fig:geneval_results}
\end{figure}

\begin{figure*}[t]
    \centering
    \includegraphics[width=\linewidth]{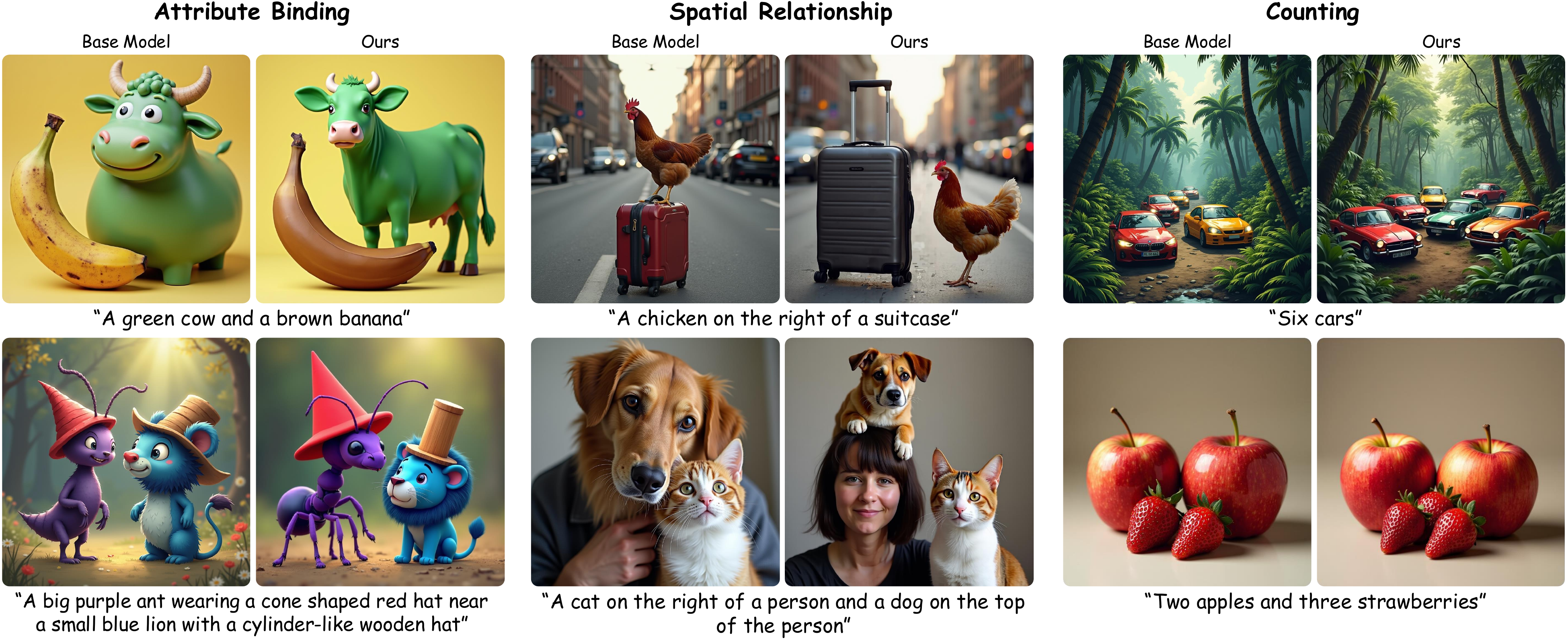}
    \vspace{-0.6cm}
    \caption{\textbf{Qualitative comparison across compositional categories.} AgentComp produces more compositionally accurate images than the base model across various categories.}
    \label{fig:compositionality_categories_qualitative}
\end{figure*}

\subsection{Agentic Generated Dataset}
In our agentic framework for data generation, we employ ChatGPT-4.1~\cite{openai2023chatgpt} as the base LLM for all agents, Stable Diffusion v3.5-Large~\cite{esser2024scaling} for image generation, Qwen-Image-Edit~\cite{wu2025qwen} for image editing, and Qwen2.5-VL-72B-Instruct~\cite{bai2025qwen2} for visual question answering (VQA). Ablation studies on alternative tool and LLM configurations are presented in Sec.~\ref{sec:ablations}. Using this setup, we generate 725 data points, each consisting of one positive image and, on average, 10 negative images, resulting in a total of approximately 9.5k images.  We employ several strategies to improve data utilization in our training pipeline. For instance, rather than relying solely on the initial positive image in each data cluster as the reference and treating the remaining images as negatives in the APO objective, we exploit additional intra-cluster relationships to construct multiple valid preference pairs. To support these strategies, we introduce auxiliary agents that handle data refinement and pair generation. Further implementation details are provided in Appendix~\ref{sec:app:experiments:dataset}, and ablations on dataset scale and the effect of human filtering are discussed in Appendix~\ref{sec:app:ablations}.

\subsection{Baselines}
We compare AgentComp with prior state-of-the-art methods on standard compositionality benchmarks. Our comparisons include a broad range of approaches: fine-tuning–based methods~\cite{huang2023t2i, han2024progressive}, LLM-planning–based methods~\cite{yang2024mastering,li2025mccd}, and inference-time optimization methods~\cite{chefer2023attend}. To evaluate model-agnostic performance, we apply AgentComp to multiple base models of varying capacities, including SDv2, SDXL~\cite{podell2023sdxl}, SDv3.5~\cite{esser2024scaling}, and FLUX~\cite{flux2024}. Across all settings, AgentComp consistently improves the compositional ability of base models and achieves state-of-the-art performance on compositionality benchmarks.

\subsection{Metrics}
To comprehensively assess the compositional performance, we employ multiple quantitative metrics. We first evaluate our method on T2I-CompBench~\cite{huang2023t2i}, a well-established benchmark that measures various aspects of compositionality, including attribute binding, spatial reasoning and relationships, counting ability, and complex compositions. This benchmark uses automatic VLM-based evaluation to compare generated images with their corresponding prompts. We further report results on GenEval~\cite{ghosh2023geneval}, another compositionality benchmark that focuses on similar aspects but with relatively simpler scenarios, as well as TIFA~\cite{hu2023tifa} benchmark (refer to Appendix~\ref{sec:app:experiments}). 

Beyond compositionality, we also evaluate overall image quality using PickScore~\cite{kirstain2023pick}, ImageReward~\cite{xu2023imagereward}, and AestheticScore~\cite{schuhmann2022laion}. Finally, since recent text-to-image models demonstrate text rendering capabilities, we include an OCR-based evaluation~\cite{liu2025flow, gong2025seedream} to assess the finetuned models’ ability to accurately generate text within images.

\subsection{Main Results}

Table~\ref{tab:model_comparison} reports AgentComp’s performance on T2I-CompBench using the FLUX-dev base model. As shown, AgentComp not only results in substantial improvements over the base model but also achieves state-of-the-art performance across nearly all categories of T2I-CompBench. Further experimental details are provided in Appendix~\ref{sec:app:experiments}. To further demonstrate its robustness and generalizability, we evaluate AgentComp with several base models of varying capacity, including SDv2, SDXL, and SDv3.5. Figure \ref{fig:compositional_baselines_improvements} shows that AgentComp consistently yields significant gains over each baseline. Finally, to comprehensively assess compositional performance beyond T2I-CompBench, we also evaluate on GenEval (Figure \ref{fig:geneval_results}), where AgentComp again demonstrates consistent improvements.

Figure~\ref{fig:compositionality_categories_qualitative} presents qualitative comparisons with the baseline, highlighting improvements across various compositional categories.
Furthermore, Figure~\ref{fig:teaser} showcases results on more complex prompts, demonstrating that AgentComp produces images aligned more faithfully with input descriptions, accurately capturing fine-grained details and complex compositions. Refer to Appendix~\ref{sec:app:experiments} for more examples.

\begin{figure}[t]
    \centering
    \includegraphics[width=\linewidth]{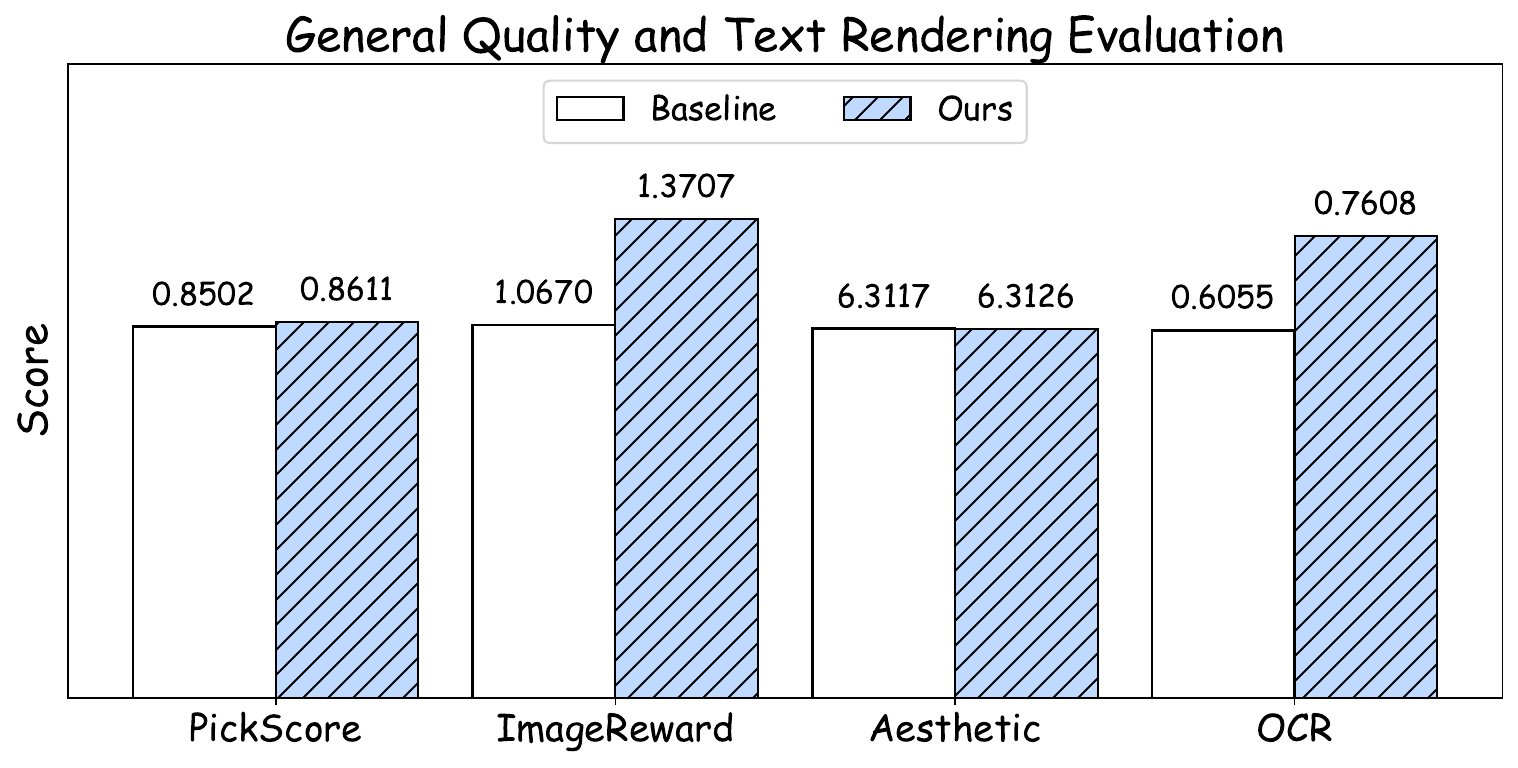}
    \vspace{-0.7cm}
    \caption{\textbf{General quality and text rendering comparison.} AgentComp preserves and even improves image quality on benchmarks while also significantly enhancing text rendering accuracy.}
    \label{fig:general_quality_text_rendering}
\end{figure}

\begin{figure}[t]
\vspace{-0.2cm}
    \centering
    \includegraphics[width=\linewidth]{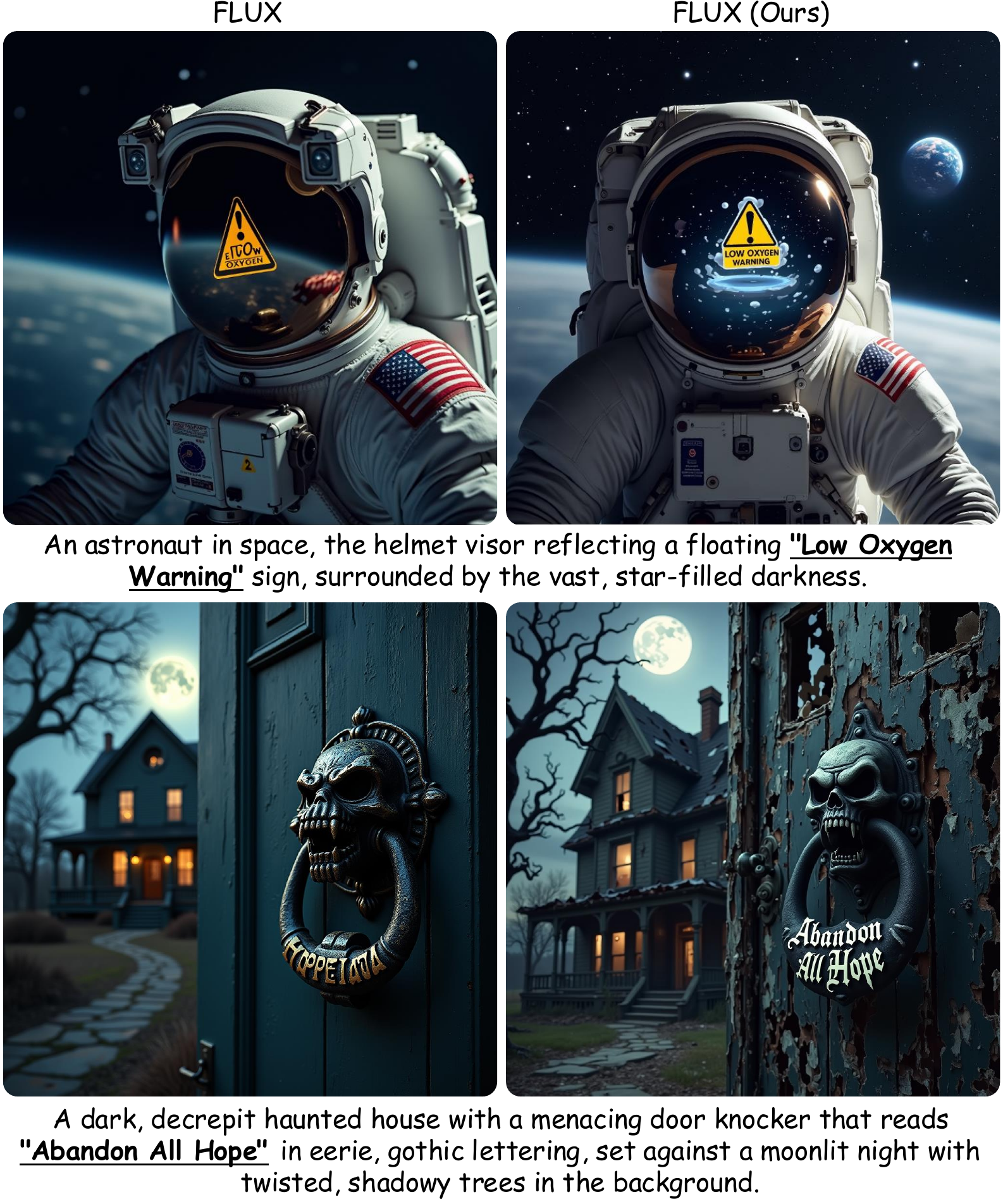}
    \vspace{-0.65cm}
    \caption{\textbf{Qualitative comparison on text rendering.} AgentComp improves text rendering accuracy and consistency in generated images (details are best viewed when zoomed in).}
    \label{fig:text_rendering_qualitative}
\end{figure}

To assess the impact of AgentComp on both the generation quality and the inherent capabilities of the base model, we evaluate it using several aesthetic and quality metrics, as well as a text-rendering benchmark. Figure \ref{fig:general_quality_text_rendering} compares the base model with AgentComp. Notably, AgentComp not only preserves the base model’s generation quality—a common drawback in prior methods—but in fact slightly improves it on some metrics, likely due to its stronger adherence to textual details in prompts. Interestingly, in the text-rendering task—despite never being explicitly trained on any such samples—AgentComp achieves significant improvements. We hypothesize that this stems from its training objective, which encourages the model to attend more closely to every detail described in the prompt. As a result, it produces images that more faithfully include textual elements that the base model often omits.
Figure \ref{fig:text_rendering_qualitative} further illustrates these qualitative gains, showing improved prompt alignment and markedly better text-rendering capabilities.

\subsection{Ablations}
\label{sec:ablations}

\textbf{Agent Ablation.} 
We perform an ablation study on the Image Generation Agent $\mathcal{A}_\text{ImageGen}$ by varying its base LLM and tools. In these experiments, the agent is evaluated on a small subset of T2I-CompBench. In the default configuration, the agent uses GPT-4.1 as the LLM, Qwen-VL-32B for VQA, SDXL for image generation, and Qwen-Image-Edit for editing. For each component, we replace it with alternative models while keeping the others fixed to assess its individual impact. Table \ref{tab:agent_ablation} summarizes the results. The findings highlight the importance of using sufficiently powerful tools and the base LLM which acts as the planner and must reason across multiple steps. Notably, smaller LLMs (e.g., GPT-OSS-20B) often fail to reason effectively and do not produce a final valid image. For fair comparison, in such cases we manually selected the last intermediate image they generated; otherwise, their performance would have been substantially lower. For additional details, including ablations on tool-calling capabilities and other statistics, refer to Appendix~\ref{sec:app:ablations}. 

 We also conduct extensive additional ablations, including analyses of the distance function used in APO, adjustments to the classifier-guidance scale after APO optimization for models that rely on explicit guidance, comparisons between APO, standard fine-tuning, and alternative fine-tuning strategies, the effect of human filtering, and inspection of attention maps, where we observe notably better alignment in AgentComp. Please refer to Appendix~\ref{sec:app:ablations} for further details and results.

\begin{table}[t]
\centering
\footnotesize
\caption{\textbf{Ablation on $\mathcal{A}_\text{ImageGen}$.} Impact of varying the base LLM and tools on compositional generation performance.}
\vspace{-0.15cm}
\setlength{\tabcolsep}{2pt}
\renewcommand{\arraystretch}{1}

\begin{tabular}{ccccccc}
\toprule
\multicolumn{4}{c}{\textbf{Agent Configuration}} & \multicolumn{3}{c}{$\textbf{T2I-CompBench}_\textbf{small}$} \\
\cmidrule(lr){1-4} \cmidrule(lr){5-7}
\textbf{LLM} 
& \textbf{VQA} 
& \textbf{ImgGen} 
& \textbf{ImgEdit} 
& \textbf{Color} 
& \textbf{Num} 
& \textbf{Spatial} \\
\midrule

\multicolumn{7}{l}{\textit{\textbf{Base Agent}}} \\
GPT-4.1 
& $\text{Qwen}_\text{32B}$
& SDXL 
& $\text{Qwen}_\text{ImgEdit}$
& 0.6895 & 0.5789 & 0.3013  \\

\midrule
\multicolumn{7}{l}{\textit{\textbf{VQA ablation}}} \\

GPT-4.1 
& \cellcolor{gray!15} $\text{Qwen}_\text{7B}$ 
& SDXL 
& $\text{Qwen}_\text{ImgEdit}$ 
& 0.6783 & 0.5693 & 0.2867  \\

GPT-4.1 
& \cellcolor{gray!15} $\text{Qwen}_\text{72B}$ 
& SDXL 
& $\text{Qwen}_\text{ImgEdit}$
& 0.7155 & 0.5839 & 0.3170  \\

\midrule
\multicolumn{7}{l}{\textit{\textbf{Image generation ablation}}} \\

GPT-4.1 
& $\text{Qwen}_\text{32B}$ 
& \cellcolor{gray!15} SDv2 
& $\text{Qwen}_\text{ImgEdit}$
& 0.6717 & 0.5543 & 0.2365   \\

GPT-4.1 
& $\text{Qwen}_\text{32B}$ 
& \cellcolor{gray!15} SDv3.5 
& $\text{Qwen}_\text{ImgEdit}$ 
& 0.8131 & 0.6796 & 0.3503  \\

\midrule
\multicolumn{7}{l}{\textit{\textbf{Image editing ablation}}} \\

GPT-4.1 
& $\text{Qwen}_\text{32B}$ 
& SDXL 
& \cellcolor{gray!15} $\text{FLUX}_\text{Kontext}$ 
& 0.7679 & 0.6066 & 0.3442  \\

\midrule
\multicolumn{7}{l}{\textit{\textbf{LLM ablation}}} \\

 \cellcolor{gray!15} $\text{gpt-oss}_\text{20B}$ 
& $\text{Qwen}_\text{32B}$ 
& SDXL 
& $\text{Qwen}_\text{ImgEdit}$
& 0.5836 & 0.5165 & 0.2117  \\

 \cellcolor{gray!15} $\text{gpt-oss}_\text{120B}$ 
& $\text{Qwen}_\text{32B}$ 
& SDXL 
& $\text{Qwen}_\text{ImgEdit}$
& 0.6532 & 0.5465 & 0.2790  \\

\bottomrule
\end{tabular}
\label{tab:agent_ablation}
\end{table}

\section{Conclusion}
We introduced AgentComp, an agentic framework that uses multi-agent orchestration to generate high-quality contrastive datasets capturing compositional variations. Through our Agent Preference Optimization objective, the model is explicitly trained to distinguish between compositionally similar generation paths and follow the correct trajectory. AgentComp achieves state-of-the-art results on compositional benchmarks while also improving image quality and text rendering, establishing a strong step toward more compositionally grounded text-to-image generation.
{
    \small
    \bibliographystyle{ieeenat_fullname}
    \bibliography{main}

@String(TOG= {ACM Trans. Graph.})

@String(ICLR = {Int. Conf. Learn. Represent.})

@String(TOG   = {ACM TOG})

@String(ICLR  = {ICLR})

@inproceedings{esser2024scaling,
  title={Scaling rectified flow transformers for high-resolution image synthesis},
  author={Esser, Patrick and Kulal, Sumith and Blattmann, Andreas and Entezari, Rahim and M{\"u}ller, Jonas and Saini, Harry and Levi, Yam and Lorenz, Dominik and Sauer, Axel and Boesel, Frederic and others},
  booktitle={Forty-first international conference on machine learning},
  year={2024}
}

@article{wu2025qwen,
  title={Qwen-image technical report},
  author={Wu, Chenfei and Li, Jiahao and Zhou, Jingren and Lin, Junyang and Gao, Kaiyuan and Yan, Kun and Yin, Sheng-ming and Bai, Shuai and Xu, Xiao and Chen, Yilei and others},
  journal={arXiv preprint arXiv:2508.02324},
  year={2025}
}

@misc{flux2024,
    author={Black Forest Labs},
    title={FLUX},
    year={2024},
    howpublished={\url{https://github.com/black-forest-labs/flux}},
}

@inproceedings{yang2024mastering,
  title={Mastering text-to-image diffusion: Recaptioning, planning, and generating with multimodal llms},
  author={Yang, Ling and Yu, Zhaochen and Meng, Chenlin and Xu, Minkai and Ermon, Stefano and Cui, Bin},
  booktitle={Forty-first International Conference on Machine Learning},
  year={2024}
}

@article{jiang2025t2i,
  title={T2i-r1: Reinforcing image generation with collaborative semantic-level and token-level cot},
  author={Jiang, Dongzhi and Guo, Ziyu and Zhang, Renrui and Zong, Zhuofan and Li, Hao and Zhuo, Le and Yan, Shilin and Heng, Pheng-Ann and Li, Hongsheng},
  journal={arXiv preprint arXiv:2505.00703},
  year={2025}
}

@article{yang2023improving,
  title={Improving diffusion-based image synthesis with context prediction},
  author={Yang, Ling and Liu, Jingwei and Hong, Shenda and Zhang, Zhilong and Huang, Zhilin and Cai, Zheming and Zhang, Wentao and Cui, Bin},
  journal={Advances in Neural Information Processing Systems},
  volume={36},
  pages={37636--37656},
  year={2023}
}

@article{chen2023pixart,
  title={Pixart-$\alpha$: Fast training of diffusion transformer for photorealistic text-to-image synthesis},
  author={Chen, Junsong and Yu, Jincheng and Ge, Chongjian and Yao, Lewei and Xie, Enze and Wu, Yue and Wang, Zhongdao and Kwok, James and Luo, Ping and Lu, Huchuan and others},
  journal={arXiv preprint arXiv:2310.00426},
  year={2023}
}

@article{podell2023sdxl,
  title={Sdxl: Improving latent diffusion models for high-resolution image synthesis},
  author={Podell, Dustin and English, Zion and Lacey, Kyle and Blattmann, Andreas and Dockhorn, Tim and M{\"u}ller, Jonas and Penna, Joe and Rombach, Robin},
  journal={arXiv preprint arXiv:2307.01952},
  year={2023}
}

@inproceedings{ramesh2021zero,
  title={Zero-shot text-to-image generation},
  author={Ramesh, Aditya and Pavlov, Mikhail and Goh, Gabriel and Gray, Scott and Voss, Chelsea and Radford, Alec and Chen, Mark and Sutskever, Ilya},
  booktitle={International conference on machine learning},
  pages={8821--8831},
  year={2021},
  organization={Pmlr}
}

@article{han2024progressive,
  title={Progressive compositionality in text-to-image generative models},
  author={Han, Evans Xu and Jin, Linghao and Liu, Xiaofeng and Liang, Paul Pu},
  journal={arXiv preprint arXiv:2410.16719},
  year={2024}
}

@inproceedings{li2025mccd,
  title={MCCD: Multi-Agent Collaboration-based Compositional Diffusion for Complex Text-to-Image Generation},
  author={Li, Mingcheng and Hou, Xiaolu and Liu, Ziyang and Yang, Dingkang and Qian, Ziyun and Chen, Jiawei and Wei, Jinjie and Jiang, Yue and Xu, Qingyao and Zhang, Lihua},
  booktitle={Proceedings of the Computer Vision and Pattern Recognition Conference},
  pages={13263--13272},
  year={2025}
}

@article{huang2023t2i,
  title={T2i-compbench: A comprehensive benchmark for open-world compositional text-to-image generation},
  author={Huang, Kaiyi and Sun, Kaiyue and Xie, Enze and Li, Zhenguo and Liu, Xihui},
  journal={Advances in Neural Information Processing Systems},
  volume={36},
  pages={78723--78747},
  year={2023}
}

@article{chefer2023attend,
  title={Attend-and-excite: Attention-based semantic guidance for text-to-image diffusion models},
  author={Chefer, Hila and Alaluf, Yuval and Vinker, Yael and Wolf, Lior and Cohen-Or, Daniel},
  journal={ACM transactions on Graphics (TOG)},
  volume={42},
  number={4},
  pages={1--10},
  year={2023},
  publisher={ACM New York, NY, USA}
}

@article{feng2022training,
  title={Training-free structured diffusion guidance for compositional text-to-image synthesis},
  author={Feng, Weixi and He, Xuehai and Fu, Tsu-Jui and Jampani, Varun and Akula, Arjun and Narayana, Pradyumna and Basu, Sugato and Wang, Xin Eric and Wang, William Yang},
  journal={arXiv preprint arXiv:2212.05032},
  year={2022}
}

@inproceedings{liu2022compositional,
  title={Compositional visual generation with composable diffusion models},
  author={Liu, Nan and Li, Shuang and Du, Yilun and Torralba, Antonio and Tenenbaum, Joshua B},
  booktitle={European conference on computer vision},
  pages={423--439},
  year={2022},
  organization={Springer}
}

@article{rafailov2023direct,
  title={Direct preference optimization: Your language model is secretly a reward model},
  author={Rafailov, Rafael and Sharma, Archit and Mitchell, Eric and Manning, Christopher D and Ermon, Stefano and Finn, Chelsea},
  journal={Advances in neural information processing systems},
  volume={36},
  pages={53728--53741},
  year={2023}
}

@inproceedings{wallace2024diffusion,
  title={Diffusion model alignment using direct preference optimization},
  author={Wallace, Bram and Dang, Meihua and Rafailov, Rafael and Zhou, Linqi and Lou, Aaron and Purushwalkam, Senthil and Ermon, Stefano and Xiong, Caiming and Joty, Shafiq and Naik, Nikhil},
  booktitle={Proceedings of the IEEE/CVF Conference on Computer Vision and Pattern Recognition},
  pages={8228--8238},
  year={2024}
}

@inproceedings{meral2024conform,
  title={Conform: Contrast is all you need for high-fidelity text-to-image diffusion models},
  author={Meral, Tuna Han Salih and Simsar, Enis and Tombari, Federico and Yanardag, Pinar},
  booktitle={Proceedings of the IEEE/CVF Conference on Computer Vision and Pattern Recognition},
  pages={9005--9014},
  year={2024}
}

@article{rassin2023linguistic,
  title={Linguistic binding in diffusion models: Enhancing attribute correspondence through attention map alignment},
  author={Rassin, Royi and Hirsch, Eran and Glickman, Daniel and Ravfogel, Shauli and Goldberg, Yoav and Chechik, Gal},
  journal={Advances in Neural Information Processing Systems},
  volume={36},
  pages={3536--3559},
  year={2023}
}

@inproceedings{bao2024separate,
  title={Separate-and-enhance: Compositional finetuning for text-to-image diffusion models},
  author={Bao, Zhipeng and Li, Yijun and Singh, Krishna Kumar and Wang, Yu-Xiong and Hebert, Martial},
  booktitle={ACM SIGGRAPH 2024 Conference Papers},
  pages={1--10},
  year={2024}
}

@article{jiang2024comat,
  title={Comat: Aligning text-to-image diffusion model with image-to-text concept matching},
  author={Jiang, Dongzhi and Song, Guanglu and Wu, Xiaoshi and Zhang, Renrui and Shen, Dazhong and Zong, Zhuofan and Liu, Yu and Li, Hongsheng},
  journal={Advances in Neural Information Processing Systems},
  volume={37},
  pages={76177--76209},
  year={2024}
}

@article{zhang2024realcompo,
  title={Realcompo: Balancing realism and compositionality improves text-to-image diffusion models},
  author={Zhang, Xinchen and Yang, Ling and Cai, Yaqi and Yu, Zhaochen and Wang, Kai-Ni and Tian, Ye and Xu, Minkai and Tang, Yong and Yang, Yujiu and Cui, Bin and others},
  journal={Advances in Neural Information Processing Systems},
  volume={37},
  pages={96963--96992},
  year={2024}
}

@article{liu2024draw,
  title={Draw like an artist: Complex scene generation with diffusion model via composition, painting, and retouching},
  author={Liu, Minghao and Zhang, Le and Tian, Yingjie and Qu, Xiaochao and Liu, Luoqi and Liu, Ting},
  journal={arXiv preprint arXiv:2408.13858},
  year={2024}
}

@inproceedings{li2023gligen,
  title={Gligen: Open-set grounded text-to-image generation},
  author={Li, Yuheng and Liu, Haotian and Wu, Qingyang and Mu, Fangzhou and Yang, Jianwei and Gao, Jianfeng and Li, Chunyuan and Lee, Yong Jae},
  booktitle={Proceedings of the IEEE/CVF conference on computer vision and pattern recognition},
  pages={22511--22521},
  year={2023}
}

@article{zhang2024itercomp,
  title={Itercomp: Iterative composition-aware feedback learning from model gallery for text-to-image generation},
  author={Zhang, Xinchen and Yang, Ling and Li, Guohao and Cai, Yaqi and Xie, Jiake and Tang, Yong and Yang, Yujiu and Wang, Mengdi and Cui, Bin},
  journal={arXiv preprint arXiv:2410.07171},
  year={2024}
}

@article{li2024all,
  title={All Seeds Are Not Equal: Enhancing Compositional Text-to-Image Generation with Reliable Random Seeds},
  author={Li, Shuangqi and Le, Hieu and Xu, Jingyi and Salzmann, Mathieu},
  journal={arXiv preprint arXiv:2411.18810},
  year={2024}
}

@article{zarei2024improving,
  title={Improving Compositional Attribute Binding in Text-to-Image Generative Models via Enhanced Text Embeddings},
  author={Zarei, Arman and Rezaei, Keivan and Basu, Samyadeep and Saberi, Mehrdad and Moayeri, Mazda and Kattakinda, Priyatham and Feizi, Soheil},
  journal={arXiv preprint arXiv:2406.07844},
  year={2024}
}

@article{zarei2024understanding,
  title={Understanding and mitigating compositional issues in text-to-image generative models},
  author={Zarei, Arman and Rezaei, Keivan and Basu, Samyadeep and Saberi, Mehrdad and Moayeri, Mazda and Kattakinda, Priyatham and Feizi, Soheil},
  journal={arXiv e-prints},
  pages={arXiv--2406},
  year={2024}
}

@article{ho2020denoising,
  title={Denoising diffusion probabilistic models},
  author={Ho, Jonathan and Jain, Ajay and Abbeel, Pieter},
  journal={Advances in neural information processing systems},
  volume={33},
  pages={6840--6851},
  year={2020}
}

@inproceedings{rombach2022high,
  title={High-resolution image synthesis with latent diffusion models},
  author={Rombach, Robin and Blattmann, Andreas and Lorenz, Dominik and Esser, Patrick and Ommer, Bj{\"o}rn},
  booktitle={Proceedings of the IEEE/CVF conference on computer vision and pattern recognition},
  pages={10684--10695},
  year={2022}
}

@inproceedings{nichol2021improved,
  title={Improved denoising diffusion probabilistic models},
  author={Nichol, Alexander Quinn and Dhariwal, Prafulla},
  booktitle={International conference on machine learning},
  pages={8162--8171},
  year={2021},
  organization={PMLR}
}

@article{song2020score,
  title={Score-based generative modeling through stochastic differential equations},
  author={Song, Yang and Sohl-Dickstein, Jascha and Kingma, Diederik P and Kumar, Abhishek and Ermon, Stefano and Poole, Ben},
  journal={arXiv preprint arXiv:2011.13456},
  year={2020}
}

@article{lipman2022flow,
  title={Flow matching for generative modeling},
  author={Lipman, Yaron and Chen, Ricky TQ and Ben-Hamu, Heli and Nickel, Maximilian and Le, Matt},
  journal={arXiv preprint arXiv:2210.02747},
  year={2022}
}

@article{labs2025flux,
  title={FLUX. 1 Kontext: Flow Matching for In-Context Image Generation and Editing in Latent Space},
  author={Labs, Black Forest and Batifol, Stephen and Blattmann, Andreas and Boesel, Frederic and Consul, Saksham and Diagne, Cyril and Dockhorn, Tim and English, Jack and English, Zion and Esser, Patrick and others},
  journal={arXiv preprint arXiv:2506.15742},
  year={2025}
}

@misc{openai2023chatgpt,
  author       = {OpenAI},
  title        = {ChatGPT (Nov 6 version)},
  year         = {2023},
  howpublished = {\url{https://chat.openai.com/}},
  note         = {Large language model},
}

@article{bai2025qwen2,
  title={Qwen2. 5-vl technical report},
  author={Bai, Shuai and Chen, Keqin and Liu, Xuejing and Wang, Jialin and Ge, Wenbin and Song, Sibo and Dang, Kai and Wang, Peng and Wang, Shijie and Tang, Jun and others},
  journal={arXiv preprint arXiv:2502.13923},
  year={2025}
}

@article{ghosh2023geneval,
  title={Geneval: An object-focused framework for evaluating text-to-image alignment},
  author={Ghosh, Dhruba and Hajishirzi, Hannaneh and Schmidt, Ludwig},
  journal={Advances in Neural Information Processing Systems},
  volume={36},
  pages={52132--52152},
  year={2023}
}

@article{kirstain2023pick,
  title={Pick-a-pic: An open dataset of user preferences for text-to-image generation},
  author={Kirstain, Yuval and Polyak, Adam and Singer, Uriel and Matiana, Shahbuland and Penna, Joe and Levy, Omer},
  journal={Advances in neural information processing systems},
  volume={36},
  pages={36652--36663},
  year={2023}
}

@article{xu2023imagereward,
  title={Imagereward: Learning and evaluating human preferences for text-to-image generation},
  author={Xu, Jiazheng and Liu, Xiao and Wu, Yuchen and Tong, Yuxuan and Li, Qinkai and Ding, Ming and Tang, Jie and Dong, Yuxiao},
  journal={Advances in Neural Information Processing Systems},
  volume={36},
  pages={15903--15935},
  year={2023}
}

@article{schuhmann2022laion,
  title={Laion-aesthetics},
  author={Schuhmann, Christoph and Beaumont, Romain},
  journal={LAION. AI},
  year={2022}
}

@article{liu2025flow,
  title={Flow-grpo: Training flow matching models via online rl},
  author={Liu, Jie and Liu, Gongye and Liang, Jiajun and Li, Yangguang and Liu, Jiaheng and Wang, Xintao and Wan, Pengfei and Zhang, Di and Ouyang, Wanli},
  journal={arXiv preprint arXiv:2505.05470},
  year={2025}
}

@article{gong2025seedream,
  title={Seedream 2.0: A native chinese-english bilingual image generation foundation model},
  author={Gong, Lixue and Hou, Xiaoxia and Li, Fanshi and Li, Liang and Lian, Xiaochen and Liu, Fei and Liu, Liyang and Liu, Wei and Lu, Wei and Shi, Yichun and others},
  journal={arXiv preprint arXiv:2503.07703},
  year={2025}
}

@inproceedings{hu2023tifa,
  title={Tifa: Accurate and interpretable text-to-image faithfulness evaluation with question answering},
  author={Hu, Yushi and Liu, Benlin and Kasai, Jungo and Wang, Yizhong and Ostendorf, Mari and Krishna, Ranjay and Smith, Noah A},
  booktitle={Proceedings of the IEEE/CVF International Conference on Computer Vision},
  pages={20406--20417},
  year={2023}
}

@inproceedings{yao2023react,
  title     = {{ReAct}: Synergizing Reasoning and Acting in Language Models},
  author    = {Yao, Shunyu and Zhao, Jeffrey and Yu, Dian and Du, Nan and Shafran, Izhak and Narasimhan, Karthik and Cao, Yuan},
  booktitle = {The Eleventh International Conference on Learning Representations (ICLR)},
  year      = {2023},
  url       = {https://openreview.net/forum?id=WE_vluYUL-X}
}

@misc{mitra2024agentinstructgenerativeteachingagentic,
      title={AgentInstruct: Toward Generative Teaching with Agentic Flows}, 
      author={Arindam Mitra and Luciano Del Corro and Guoqing Zheng and Shweti Mahajan and Dany Rouhana and Andres Codas and Yadong Lu and Wei-ge Chen and Olga Vrousgos and Corby Rosset and Fillipe Silva and Hamed Khanpour and Yash Lara and Ahmed Awadallah},
      year={2024},
      eprint={2407.03502},
      archivePrefix={arXiv},
      primaryClass={cs.AI},
      url={https://arxiv.org/abs/2407.03502}, 
}

@article{zarei2025localizing,
  title={Localizing Knowledge in Diffusion Transformers},
  author={Zarei, Arman and Basu, Samyadeep and Rezaei, Keivan and Lin, Zihao and Nag, Sayan and Feizi, Soheil},
  journal={arXiv preprint arXiv:2505.18832},
  year={2025}
}

@article{wei2022chain,
  title={Chain-of-thought prompting elicits reasoning in large language models},
  author={Wei, Jason and Wang, Xuezhi and Schuurmans, Dale and Bosma, Maarten and Xia, Fei and Chi, Ed and Le, Quoc V and Zhou, Denny and others},
  journal={Advances in neural information processing systems},
  volume={35},
  pages={24824--24837},
  year={2022}
}

@article{schick2023toolformer,
  title={Toolformer: Language models can teach themselves to use tools},
  author={Schick, Timo and Dwivedi-Yu, Jane and Dess{\`\i}, Roberto and Raileanu, Roberta and Lomeli, Maria and Hambro, Eric and Zettlemoyer, Luke and Cancedda, Nicola and Scialom, Thomas},
  journal={Advances in Neural Information Processing Systems},
  volume={36},
  pages={68539--68551},
  year={2023}
}

@article{patil2024gorilla,
  title={Gorilla: Large language model connected with massive apis},
  author={Patil, Shishir G and Zhang, Tianjun and Wang, Xin and Gonzalez, Joseph E},
  journal={Advances in Neural Information Processing Systems},
  volume={37},
  pages={126544--126565},
  year={2024}
}

@article{shen2023hugginggpt,
  title={Hugginggpt: Solving ai tasks with chatgpt and its friends in hugging face},
  author={Shen, Yongliang and Song, Kaitao and Tan, Xu and Li, Dongsheng and Lu, Weiming and Zhuang, Yueting},
  journal={Advances in Neural Information Processing Systems},
  volume={36},
  pages={38154--38180},
  year={2023}
}

@article{hu2024visual,
  title={Visual sketchpad: Sketching as a visual chain of thought for multimodal language models},
  author={Hu, Yushi and Shi, Weijia and Fu, Xingyu and Roth, Dan and Ostendorf, Mari and Zettlemoyer, Luke and Smith, Noah A and Krishna, Ranjay},
  journal={Advances in Neural Information Processing Systems},
  volume={37},
  pages={139348--139379},
  year={2024}
}

@article{wu2023visual,
  title={Visual chatgpt: Talking, drawing and editing with visual foundation models},
  author={Wu, Chenfei and Yin, Shengming and Qi, Weizhen and Wang, Xiaodong and Tang, Zecheng and Duan, Nan},
  journal={arXiv preprint arXiv:2303.04671},
  year={2023}
}

@article{yang2023mm,
  title={Mm-react: Prompting chatgpt for multimodal reasoning and action},
  author={Yang, Zhengyuan and Li, Linjie and Wang, Jianfeng and Lin, Kevin and Azarnasab, Ehsan and Ahmed, Faisal and Liu, Zicheng and Liu, Ce and Zeng, Michael and Wang, Lijuan},
  journal={arXiv preprint arXiv:2303.11381},
  year={2023}
}

@article{zarei2025slideredit,
  title={SliderEdit: Continuous Image Editing with Fine-Grained Instruction Control},
  author={Zarei, Arman and Basu, Samyadeep and Pournemat, Mobina and Nag, Sayan and Rossi, Ryan and Feizi, Soheil},
  journal={arXiv preprint arXiv:2511.09715},
  year={2025}
}

@article{Bradley1952RANKAO,
  title={RANK ANALYSIS OF INCOMPLETE BLOCK DESIGNS THE METHOD OF PAIRED COMPARISONS},
  author={Ralph Allan Bradley and Milton E. Terry},
  journal={Biometrika},
  year={1952},
  volume={39},
  pages={324-345},
  url={https://api.semanticscholar.org/CorpusID:121987403}
}

@inproceedings{li2022blip,
  title={Blip: Bootstrapping language-image pre-training for unified vision-language understanding and generation},
  author={Li, Junnan and Li, Dongxu and Xiong, Caiming and Hoi, Steven},
  booktitle={International conference on machine learning},
  pages={12888--12900},
  year={2022},
  organization={PMLR}
}
}

\clearpage
\setcounter{page}{1}
\maketitlesupplementary

\section{Method}
\label{sec:app:method}

\subsection{Agentic Contrastive Dataset Generation}
\label{sec:app:method:agentic_dataset_generation}
In this section, we present example instructions used within our agentic orchestra. Figure~\ref{fig:image_generation_agent_instructions} shows a sample instruction provided to the image-generation agent, while Figure~\ref{fig:image_edit_agent_instructions} illustrates instructions given to the image-editing agent. By specifying each task clearly and supplying demonstrations through in-context, few-shot examples, the agents reliably infer the intended behavior and execute the tasks with high accuracy. Furthermore, Section~\ref{sec:app:experiments:dataset} provides additional details on the techniques we employ to enhance data utilization throughout training.

\subsection{Agent Preference Optimization}
\label{sec:app:apo}

In this section, we provide a step-by-step derivation of APO.  Following the standard formulation in the preference learning literature, each sample in our dataset contains a ranked pair under a given conditioning prompt: $\mathcal{I}^+ \succ \mathcal{I}^- | \mathcal{P}$ (for some $\mathcal{I}^-$ in $\{\mathcal{I}^-_k\}_{k=1}^K$). According to the Bradley–Terry model \cite{Bradley1952RANKAO}, the probability of preferring $\mathcal{I}^+$ over $\mathcal{I}^-$ is defined as:
$$
p_\text{BT}(\mathcal{I}^+ \succ \mathcal{I}^- | \mathcal{P}) = \sigma (r(\mathcal{P}, \mathcal{I}^+) - r(\mathcal{P}, \mathcal{I^-})).
$$
where $r(\mathcal{P}, \mathcal{I})$ represents a reward function parameterized by a neural network $\psi$. The reward model can be learned via maximum-likelihood estimation:
$$
\mathcal{L}_\text{BT}(\psi) = - \mathbb{E}_{\mathcal{P},\mathcal{I}^+, \mathcal{I}^-}[\log \sigma (r_\psi(\mathcal{P}, \mathcal{I}^+) - r_\psi(\mathcal{P}, \mathcal{I}^-))]
$$
In reinforcement learning from human feedback (RLHF), the goal is to learn a conditional distribution $p_\theta(\mathcal{I}|\mathcal{P})$ that maximizes the expected reward while remaining close to a reference model $p_\text{ref}(\mathcal{I}|\mathcal{P})$. This can be expressed as:
\begin{align*}
\max_{p_\theta} \;\; & \mathbb{E}_{\mathcal{P}\sim \mathcal{D}_\mathcal{P}, \mathcal{I}\sim p_\theta(\mathcal{I}|\mathcal{P})}[r(\mathcal{P}, \mathcal{I})] \\
&- \beta \mathbb{D}_\text{KL}[p_\theta(\mathcal{I}|\mathcal{P}) \| p_\text{ref}(\mathcal{I}|\mathcal{P})]
\end{align*}

In the equation above, following \cite{rafailov2023direct}, the unique global optimal solution $p_\theta^*$ takes the form:
\begin{align*}    
p_\theta^*(\mathcal{I}|\mathcal{P}) &= p_\text{ref}(\mathcal{I}|\mathcal{P}) \exp(r(\mathcal{P}, \mathcal{I}) / \beta) / Z(\mathcal{P}) \\
Z(\mathcal{P}) &= \sum_{\mathcal{I}'} p_\text{ref}(\mathcal{I}'|\mathcal{P}) \exp(r(\mathcal{P}, \mathcal{I}') / \beta)
\end{align*}
Therefore, the reward function can be expressed as:
\[
r(\mathcal{P}, \mathcal{I}) = \beta\log\frac{p_\theta^*(\mathcal{I}|\mathcal{P})}{p_\text{ref}(\mathcal{I}|\mathcal{P})} + \beta \log Z(\mathcal{P})
\]

From the definition of $\mathcal{L}_\text{BT}$, the reward objective becomes:
\begin{equation*}
\scriptsize
\begin{aligned}
\mathcal{L}_\text{DPO}(\theta) = -\mathbb{E}_{\mathcal{P}, \mathcal{I}^+, \mathcal{I}^-} \Big[ \log \sigma \big( \beta \log\frac{p_\theta(\mathcal{I}^+ |\mathcal{P})}{p_\text{ref}(\mathcal{I}^+ |\mathcal{P})} - \beta \log\frac{p_\theta(\mathcal{I}^- |\mathcal{P})}{p_\text{ref}(\mathcal{I}^- |\mathcal{P})} \big) \Big]
\end{aligned}
\end{equation*}

With this reparameterization, rather than optimizing a reward function and subsequently applying RL, we directly optimize the optimal conditional distribution.

Following \cite{wallace2024diffusion} for adapting DPO for diffusion models, this leads to:
\begin{equation*}
\small
\begin{aligned}
&\mathcal{L}_\text{DPO-Diffusion}(\theta) = -\mathbb{E}_{\mathcal{P}, \mathcal{I}^+, \mathcal{I}^-} \log \sigma \Big( \\ &\beta \mathbb{E}_{\substack{\mathcal{I}^+_{1:T} \sim p_\theta(\mathcal{I}^+_{1:T} |\mathcal{I}^+_0) \\ \mathcal{I}^-_{1:T} \sim p_\theta(\mathcal{I}^-_{1:T} |\mathcal{I}^-_0)}} \Big[ \log\frac{p_\theta(\mathcal{I}^+_{0:T} |\mathcal{P})}{p_\text{ref}(\mathcal{I}^+_{0:T} |\mathcal{P})} - \beta \log\frac{p_\theta(\mathcal{I}^-_{0:T} |\mathcal{P})}{p_\text{ref}(\mathcal{I}^-_{0:T} |\mathcal{P})} \Big] \Big)
\end{aligned}
\end{equation*}
After some algebraic manipulation and simplification, this yields the following loss:
\begin{equation*}
\small
\begin{aligned}
\mathcal{L}(\theta) &= - \mathbb{E}_{(\mathcal{I}^+, \mathcal{I}^-, \mathcal{P}) \sim \mathcal{D}} \Big[ 
 \log \sigma \big(-\beta T \,  \, ( \ell^+ - \ell^- ) \big)  \Big] \\
\ell^+ &= \| \epsilon^+ - \epsilon_\theta(\mathcal{I}^+_t, \mathcal{P}, t) \|_2^2 - \| \epsilon^+ - \epsilon_{\text{ref}}(\mathcal{I}^+_t, \mathcal{P}, t) \|_2^2 \\
 \ell^- &= \| \epsilon^- - \epsilon_\theta(\mathcal{I}^-_t, \mathcal{P}, t) \|_2^2 - \| \epsilon^- - \epsilon_{\text{ref}}(\mathcal{I}^-_t, \mathcal{P}, t) \|_2^2 
\end{aligned}
\end{equation*}
where $\mathcal{I}^{+/-}_t$ denotes the noisy latent sample at diffusion timestep $t$, $\epsilon_\theta$ represents the trainable denoising network, $\epsilon_\text{ref}$ the reference (base) model, $\epsilon^{+/-}$ the ground-truth denoising target, and the expectation is taken over $t\sim\mathcal{U}[0,1]$.

This formulation treats all negative samples equally within a ranked group $\{\mathcal{I}^-_k\}_{k=1}^K$, regardless of their compositional proximity to the positive sample.
To better leverage the fine-grained distances estimated by our agentic orchestration, we reweight the loss by introducing a distance-dependent scaling function $\mathcal{H}(.)$ applied to $\beta$. By reweighting $\beta$, we intuitively tune how freely the model can deviate from the reference to satisfy the preference.
Our final Agent Preference Optimization (APO) objective becomes:
\vspace{5pt}
\begin{equation*}
\footnotesize
\begin{aligned}
\mathcal{L}_\text{APO}(\theta) = - \mathbb{E}_{(\mathcal{I}^+, \mathcal{I}_k^-, \mathcal{P}, d_k) \sim \mathcal{D}} \Big[ 
 \log \sigma \big(- \mathcal{H}(d_k) \beta T \,  \, ( \ell^+ - \ell^- ) \big)  \Big]
\end{aligned}
\vspace{5pt}
\end{equation*}
This distance-aware weighting offers finer control over sample importance, resulting in faster and more reliable convergence during fine-tuning.

\section{Experiments and Results}
\label{sec:app:experiments}

\subsection{Implementation Details}
\label{sec:app:implementation_details}

\begin{figure}[t]
    \centering
    \includegraphics[width=0.9\linewidth]{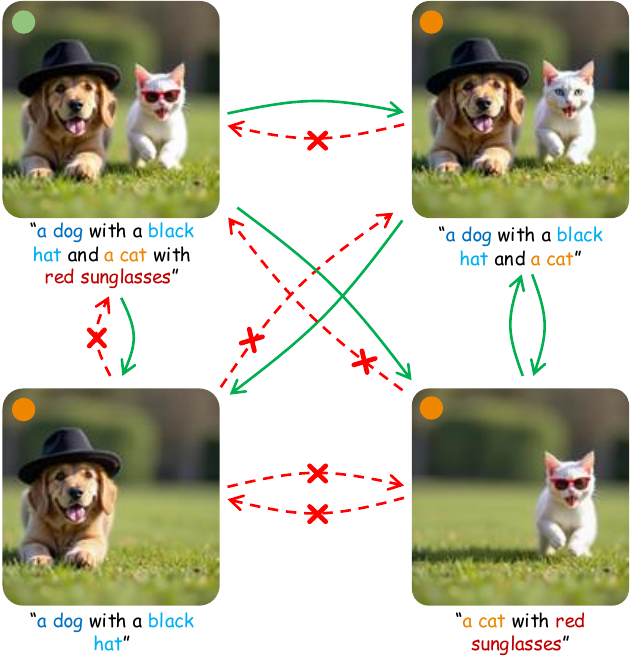}
    \vspace{-0.10cm}
    \caption{Illustration of intra-cluster contrastive pair relationships. An additional agent is introduced to identify valid contrastive pairs among samples generated for a source prompt/image (top left). Each arrow starts at the positive sample and ends at the negative. Green arrows denote valid contrastive pairs, while red arrows denote invalid ones.}
    \label{fig:contrastive_pair_filter}
\end{figure}

We use FLUX-dev as our primary base model in all experiments. To evaluate the generality and robustness of our approach, we additionally apply it to SDv3.5, SDXL, and SDv2, covering models with a broad range of capacities. All models are fine-tuned using LoRA (rank 32) for efficiency. Furthermore, consistent with prior observations in \cite{esser2024scaling, zarei2025localizing, zarei2025slideredit}, we found that training adapters on a subset of DiT blocks performs comparably to training all blocks, though we do not leverage this variant here and leave it for future investigation. For FLUX, we set the base value of $\beta = 100$ in the APO objective and employ a normalized distance function in the range $[0.5,1]$, assigning lower $\beta$ to smaller distances and higher $\beta$ to larger distances. This adaptively scales $\beta$ between 50 and 100, encouraging the model to focus more on subtle, compositionally challenging trajectories. During FLUX training, the CFG scale is fixed to 1, which we find necessary for stable optimization. Training is performed at a fixed resolution of $1024\times1024$ with a global batch size of 128 (64 positive–negative pairs), using bfloat16 mixed precision, gradient checkpointing, and gradient clipping (clip norm = 1). We optimize using Adam with a learning rate of 1e-4 and no warm-up schedule. All experiments are run on 8×H100 GPUs using distributed training. We note that larger effective batch sizes further improve performance; this can be achieved by using higher-memory GPUs, scaling to more nodes, or employing gradient accumulation. A systematic study of batch-size scaling is left for future work.

\subsection{Improving Dataset Utilization}
\label{sec:app:experiments:dataset}

In this section, we describe how we increase data utilization within our framework.  
As outlined in Section~\ref{sec:method}, the agentic orchestra produces, for each prompt--image pair $(\mathcal{P}, \mathcal{I})$, a set of contrastive samples  
\[
\{(\mathcal{P}_k^-, \mathcal{I}_k^-, d_k)\}_{k=1}^K,
\]
which form a cluster of semantically related but compositionally distinct candidates.  
A naive approach would use $(\mathcal{P}, \mathcal{I})$ as the positive sample in APO and select a single $(\mathcal{P}_k^-, \mathcal{I}_k^-)$ at random as the negative sample.  
However, this ignores the rich structure inside each cluster: many pairs \[
\big( (\mathcal{P}_{k_A}^-, \mathcal{I}_{k_A}^-),\; (\mathcal{P}_{k_B}^-, \mathcal{I}_{k_B}^-) \big)
\]
themselves form valid positive–negative relationships and can therefore be used directly for preference optimization.

Figure~\ref{fig:contrastive_pair_filter} illustrates this idea.  
The top-left image shows the original positive pair $(\mathcal{P}, \mathcal{I})$, while the remaining images display its three derived negative samples $\{(\mathcal{P}_k^-, \mathcal{I}_k^-)\}_{k=1}^3$.  
These three negatives can form additional meaningful contrastive relationships among themselves.
For instance, the sample \emph{``a dog with a black hat and a cat''} can be treated as a positive example relative to \emph{``a dog with a black hat''}, since the latter omits the information \emph{``a cat''}. 
Identifying such intra-cluster relationships allows us to construct additional reliable preference pairs beyond those involving the original $(\mathcal{P}, \mathcal{I})$.

To automatically discover these relationships, we introduce an additional agent---a \emph{contrastive pair filter agent}.  
For any candidate pair  
\[
\big( (\mathcal{P}_{k_A}^-, \mathcal{I}_{k_A}^-),\; (\mathcal{P}_{k_B}^-, \mathcal{I}_{k_B}^-) \big),
\]
the agent determines whether the second sample can serve as a valid negative for the first.  
By querying this agent across all pairwise combinations within each cluster, we significantly expand the set of usable positive–negative pairs.  
This substantially improves data utilization and enables preference optimization to exploit the full semantic structure present in the generated clusters.

\begin{figure*}[t]
    \centering
    \includegraphics[width=0.95\linewidth]{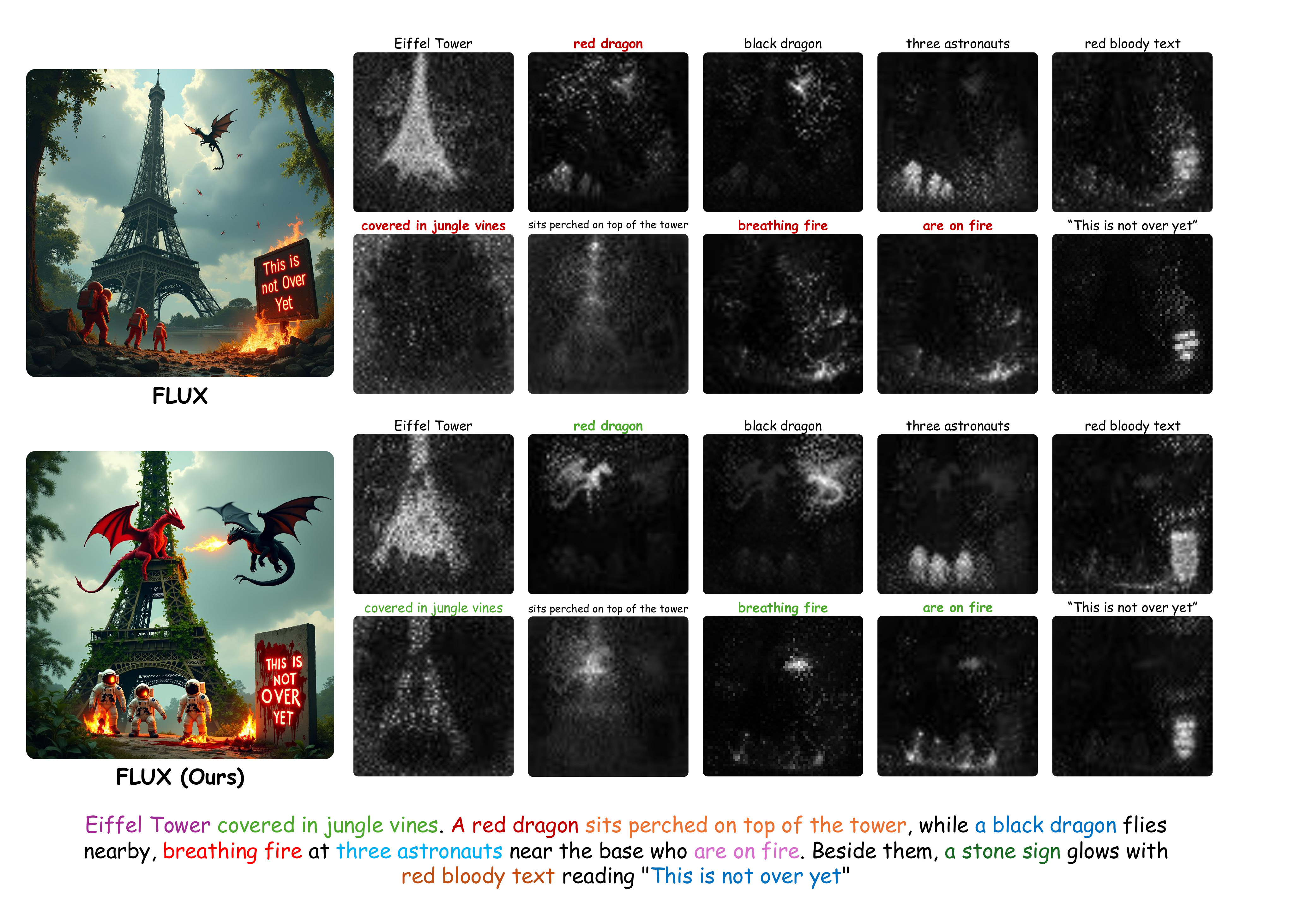}
    \vspace{-0.15cm}
    \caption{Visualization of aggregated attention maps for selected prompt tokens during denoising. The APO-trained model shows more accurate, localized attention that aligns with the intended semantics, leading to more compositionally faithful generations.}
    \label{fig:attention_map_visualization}
\end{figure*}

\subsection{Main Results}
Table~\ref{tab:model_comparison} compares AgentComp with baseline models and state-of-the-art methods for improving compositionality. For fairness and consistency, we re-ran all baseline models—SDXL, SDv2, SDv3.5, Qwen, and FLUX—under the T2I-CompBench evaluation pipeline. For prior methods, we gathered results reported across multiple papers and selected the most consistent numbers. In several cases, however, we noticed discrepancies between the original reported results and those reproduced by independent works. To ensure a fair comparison, we reproduced these methods directly using their publicly released codebases (e.g., RPG), sweeping across all hyperparameter settings recommended by the authors and selecting the best-performing result for each T2I-CompBench category. Despite this careful reproduction, we still observed substantial gaps between the results originally reported and those obtained using the official code under the documented settings.

\subsection{Text Rendering}
\begin{figure*}[t]
    \centering
    \includegraphics[width=\linewidth]{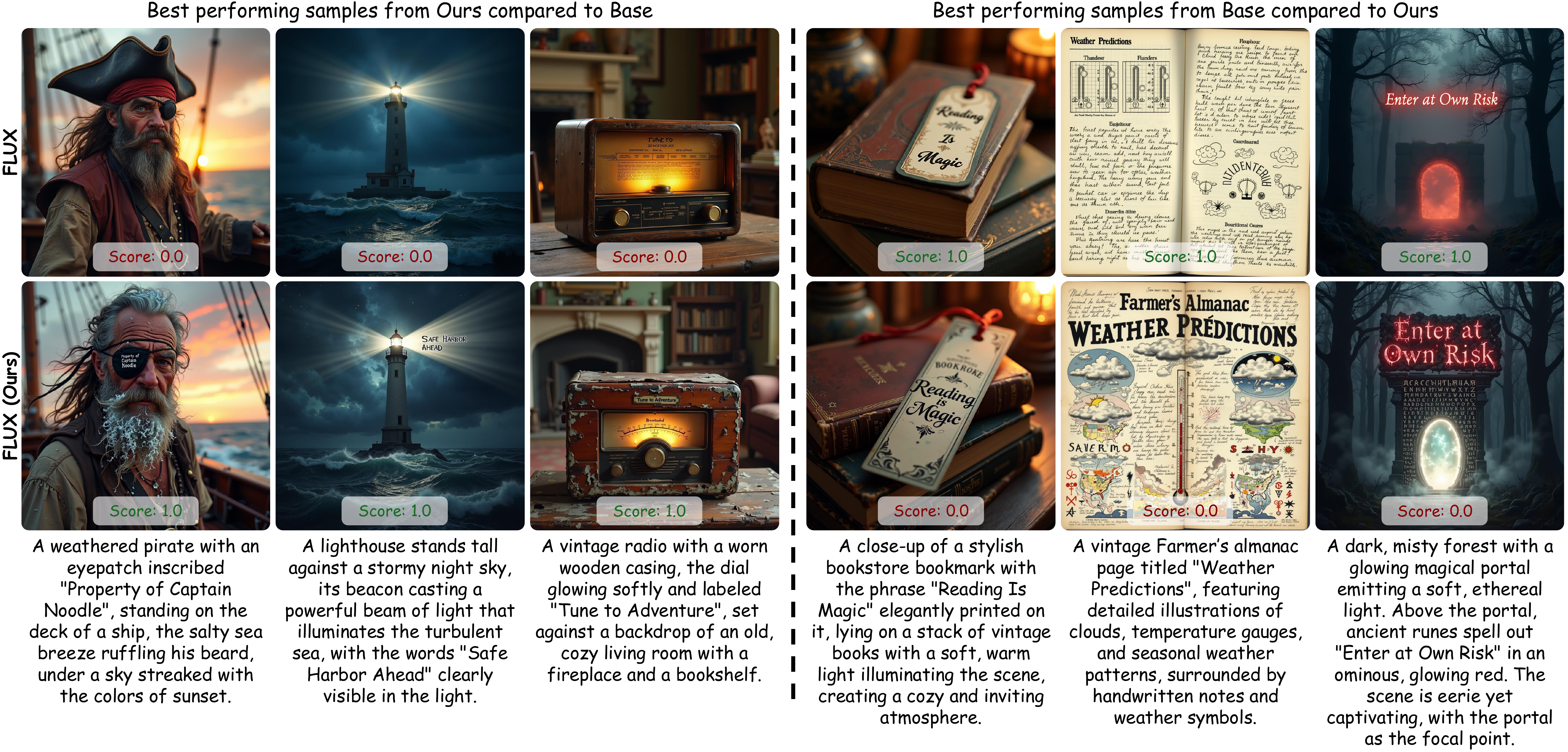}
    \vspace{-0.55cm}
    \caption{Qualitative text-generation comparison. Left panel shows cases where AgentComp succeeds while the base model fails, and the right panel shows cases where the base model performs better. Many zero-score cases for AgentComp arise from OCR detection errors rather than incorrect generation.}
    \label{fig:ocr_sorted_results}
\end{figure*}

In this section, we provide additional qualitative examples highlighting the text-rendering capabilities of FLUX and AgentComp. Figure~\ref{fig:ocr_sorted_results} presents outputs from both models on the OCR evaluation set. On the left, we show cases where AgentComp achieves high OCR scores while the base model fails; on the right, we show the opposite ordering. Interestingly, in many examples where the evaluator assigns a score of zero to AgentComp, the issue often stems from detection failures in the OCR evaluator itself rather than an actual mistake in the generated text by AgentComp.

\begin{figure}[t]
\vspace{-0.5cm}
    \centering
    \includegraphics[width=\linewidth]{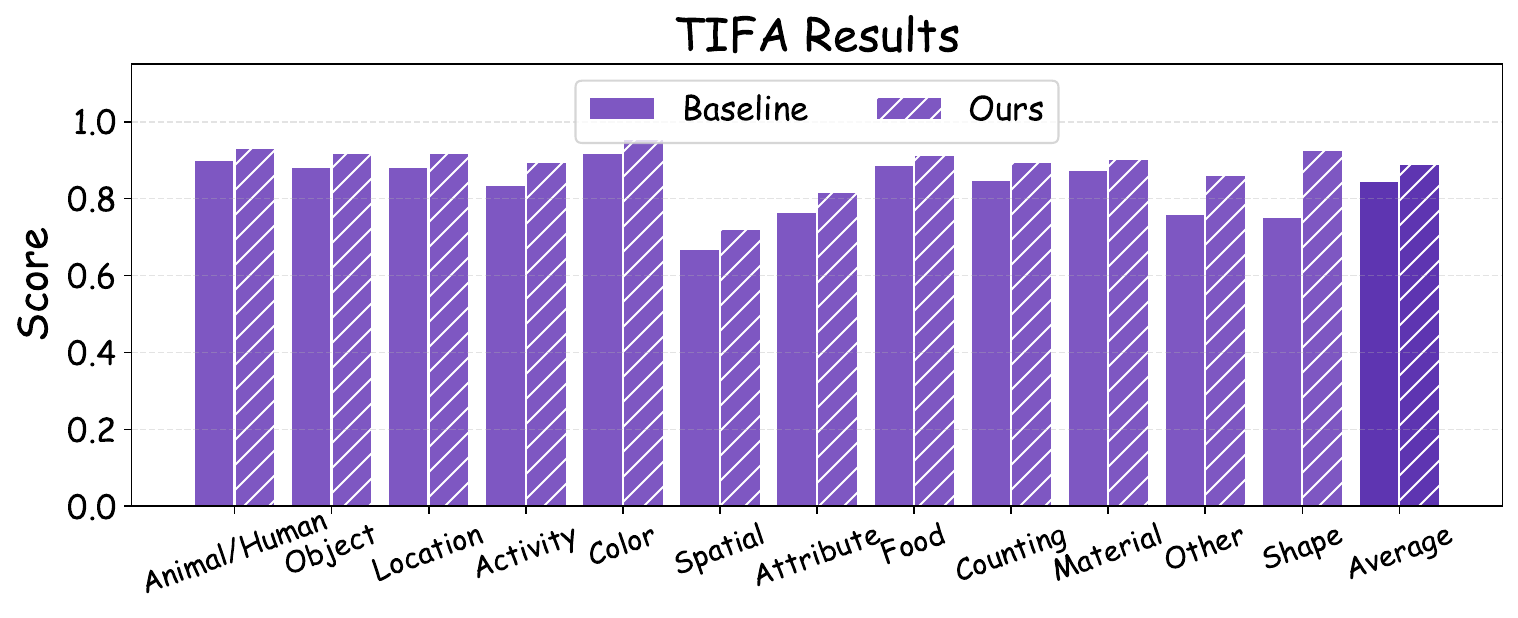}
    \vspace{-0.55cm}
    \caption{TIFA results comparing baseline and ours. AgentComp consistently improves performance across all categories.}
    \label{fig:tifa_results}
\end{figure}

\begin{figure}[t]
\vspace{-0.5cm}
    \centering
    \includegraphics[width=\linewidth]{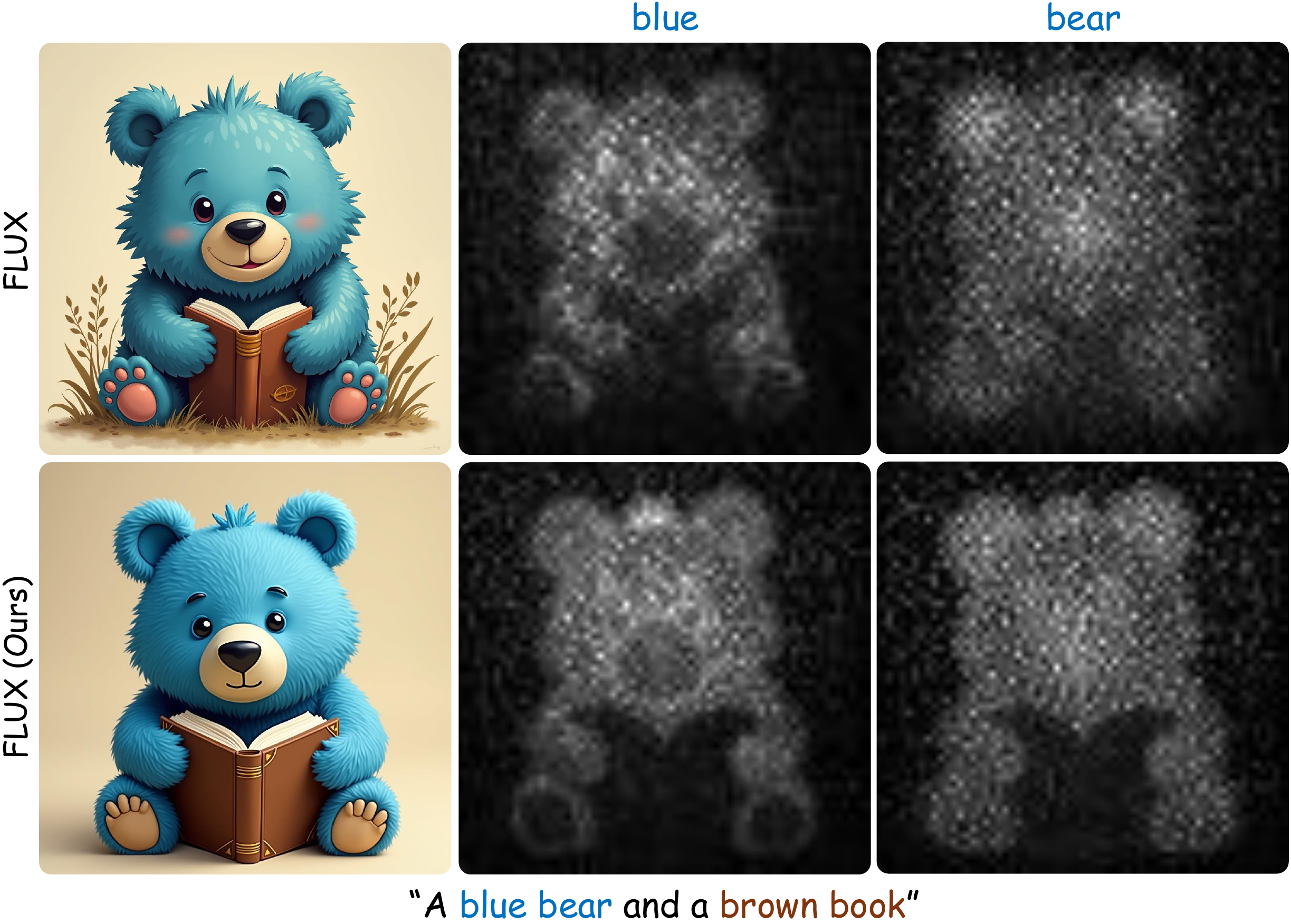}
    \vspace{-0.55cm}
    \caption{Attention map comparison for a case where both the base model and CompAgent generate correct images. CompAgent exhibits more precise, localized attention with reduced leakage into irrelevant regions.}
    \label{fig:attention_map_visualization_2}
\end{figure}

\subsection{TIFA Evaluation}
To further assess the compositional performance of AgentComp, we extend our evaluation beyond T2I-CompBench and GenEval by conducting additional experiments on the TIFA benchmark~\cite{hu2023tifa}. Using BLIP-Large~\cite{li2022blip} as the VQA model, we evaluate both FLUX and FLUX + AgentComp on this benchmark. As shown in Figure~\ref{fig:tifa_results}, AgentComp delivers consistent improvements across all TIFA categories, demonstrating the robustness and effectiveness of our approach.

\subsection{Ablations}
\label{sec:app:ablations}

\subsubsection{Accurate Attention Maps}

To interpret how our APO objective improves the model’s compositional understanding—such as attribute binding—we visualize the model’s internal attention behavior during denoising. Using FLUX as the base model, we extract the attention maps associated with a chosen group of prompt tokens at each denoising step, then aggregate these maps across all transformer blocks and timesteps. This provides an intuitive picture of where the model attends when grounding specific textual semantics in the image. Figure~\ref{fig:attention_map_visualization} presents an example. As shown, our APO-trained model produces much more accurate and localized attention patterns that more faithfully correspond to the intended prompt details, leading to generally more compositionally accurate generations. 
Moreover, Figure~\ref{fig:attention_map_visualization_2} illustrates a case where both the base model and CompAgent produce correct images; however, the CompAgent variant exhibits sharper, more precisely localized attention. Its token-wise attention shows fewer leaks into irrelevant regions, better aligning with the intended semantic areas.

\subsubsection{Fine-tuning Strategies Ablation}
\begin{table*}[t]
\centering
\footnotesize
\caption{\textbf{Ablation on $\mathcal{A}_\text{ImageGen}$.} Impact of varying the base LLM and tools on compositional generation and tool calling performance.}
\vspace{-0.15cm}
\setlength{\tabcolsep}{4pt}
\renewcommand{\arraystretch}{1}

\begin{tabular}{cccccccccccc}
\toprule
\multicolumn{4}{c}{\textbf{Agent Configuration}} & \multicolumn{3}{c}{$\textbf{T2I-CompBench}_\textbf{small}$} & \multicolumn{3}{c}{\textbf{\# of Tool Calls}} & \multirow{3}{*}{\parbox{2cm}{\centering\textbf{\# of\\Output Images}}} \\
\cmidrule(lr){1-4} \cmidrule(lr){5-7} \cmidrule(lr){8-10} 
\textbf{LLM} 
& \textbf{VQA} 
& \textbf{ImgGen} 
& \textbf{ImgEdit} 
& \textbf{Color} 
& \textbf{Num} 
& \textbf{Spatial} 
& \textbf{VQA}
& \textbf{ImageGen}
& \textbf{ImageEdit}
&  \\
\midrule

\multicolumn{11}{l}{\textit{\textbf{Base Agent}}} \\
GPT-4.1 
& $\text{Qwen}_\text{32B}$
& SDXL 
& $\text{Qwen}_\text{ImgEdit}$
& 0.6895 & 0.5789 & 0.3013 & 3328 & 763 & 2572 & 544 / 895 \\

\midrule
\multicolumn{11}{l}{\textit{\textbf{VQA ablation}}} \\

GPT-4.1 
& \cellcolor{gray!15} $\text{Qwen}_\text{7B}$ 
& SDXL 
& $\text{Qwen}_\text{ImgEdit}$ 
& 0.6783 & 0.5693 & 0.2867 & 3478 & 898 & 2573 & 547 / 895 \\

GPT-4.1 
& \cellcolor{gray!15} $\text{Qwen}_\text{72B}$ 
& SDXL 
& $\text{Qwen}_\text{ImgEdit}$
& 0.7155 & 0.5839 & 0.3170 & 4342 & 1817 & 2550 & 425 / 895 \\

\midrule
\multicolumn{11}{l}{\textit{\textbf{Image generation ablation}}} \\

GPT-4.1 
& $\text{Qwen}_\text{32B}$ 
& \cellcolor{gray!15} SDv2 
& $\text{Qwen}_\text{ImgEdit}$
& 0.6717 & 0.5543 & 0.2365 & 3354 & 808 & 2550 & 550 / 895 \\

GPT-4.1 
& $\text{Qwen}_\text{32B}$ 
& \cellcolor{gray!15} SDv3.5 
& $\text{Qwen}_\text{ImgEdit}$ 
& 0.8131 & 0.6796 & 0.3503 & 3002 & 545 & 2466 & 713 / 895 \\

\midrule
\multicolumn{11}{l}{\textit{\textbf{Image editing ablation}}} \\

GPT-4.1 
& $\text{Qwen}_\text{32B}$ 
& SDXL 
& \cellcolor{gray!15} $\text{FLUX}_\text{Kontext}$ 
& 0.7679 & 0.6066 & 0.3442 & 4252 & 1732 & 2535 & 494 / 895 \\

\midrule
\multicolumn{11}{l}{\textit{\textbf{LLM ablation}}} \\

 \cellcolor{gray!15} $\text{gpt-oss}_\text{20B}$ 
& $\text{Qwen}_\text{32B}$ 
& SDXL 
& $\text{Qwen}_\text{ImgEdit}$
& 0.5836 & 0.5165 & 0.2117 & 589 & 225 & 1044 & 107 / 895 \\

 \cellcolor{gray!15} $\text{gpt-oss}_\text{120B}$ 
& $\text{Qwen}_\text{32B}$ 
& SDXL 
& $\text{Qwen}_\text{ImgEdit}$
& 0.6532 & 0.5465 & 0.2790 & 1905 & 1197 & 2666 & 186 / 895 \\

\bottomrule
\end{tabular}
\label{tab:agent_ablation_stats_appendix}
\end{table*}

In this section, we design a small-scale experiment to compare different fine-tuning strategies and demonstrate the effectiveness of our APO objective. We evaluate four approaches: (1) standard diffusion fine-tuning, (2) a batch-variant of diffusion fine-tuning in which samples from the same cluster ($\mathcal{I}^+$ along with several $\mathcal{I}^-_k$) are presented together to help the model better distinguish subtle differences, (3) diffusion DPO, and (4) our APO method, which introduces a dynamic weighting mechanism that allows the model to deviate from the reference model when necessary to avoid compositionally incorrect trajectories.

\begin{figure}[t]
\vspace{-0.3cm}
    \centering
    \includegraphics[width=\linewidth]{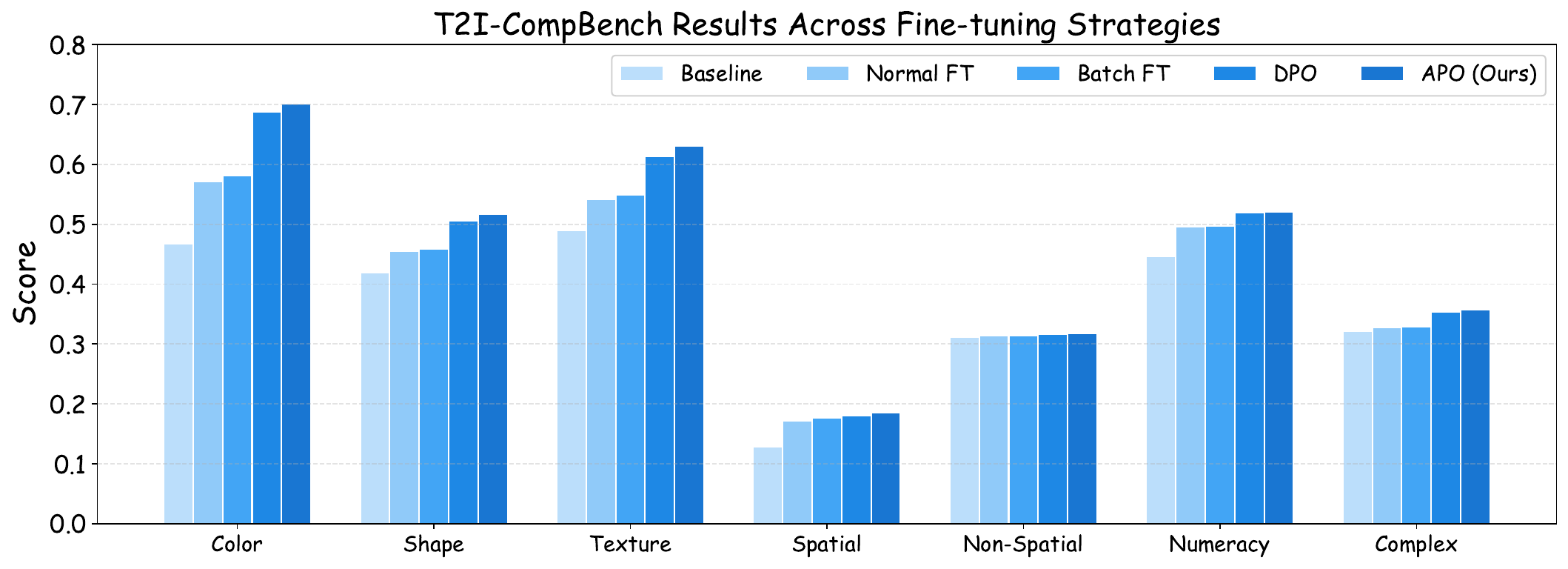}
    \vspace{-0.55cm}
    \caption{Comparison of fine-tuning strategies. APO outperforms standard diffusion fine-tuning, its batch-variant, and diffusion DPO by dynamically adjusting the model’s flexibility to correct compositionally incorrect trajectories.}
    \label{fig:ft_strategies_comparison}
\end{figure}

All models are fine-tuned under identical settings using SDv2. The results, summarized in Figure~\ref{fig:ft_strategies_comparison}, show that while the batch-enhanced diffusion baseline offers a slight improvement, both methods remain far behind DPO and APO. Moreover, although DPO already provides a substantial boost over the baselines, APO achieves further gains by dynamically adjusting the model’s flexibility to stay close to the reference model or deviate when correcting high-impact compositional errors during preference optimization.

\subsubsection{APO Distance Function Ablation}
For the weighting function $\mathcal{H}(\cdot)$ used in our APO objective, we experimented with several functional forms and weighting schemes. We initially evaluated both exponential and linear schedules that increase or decrease with respect to the pairwise distance. Since these alternatives did not yield meaningful performance differences, we adopted the linear schedule for simplicity.

We also explored applying the weighting function outside the sigmoid term in APO—effectively reweighting the loss after the fact—but found that directly scaling the parameter $\beta$ produces consistently better results. In addition, we tested two opposite strategies for mapping distances to $\beta$: assigning smaller $\beta$ to higher distances, and assigning smaller $\beta$ to lower distances.

\begin{figure}[t]
\vspace{-0.3cm}
    \centering
    \includegraphics[width=\linewidth]{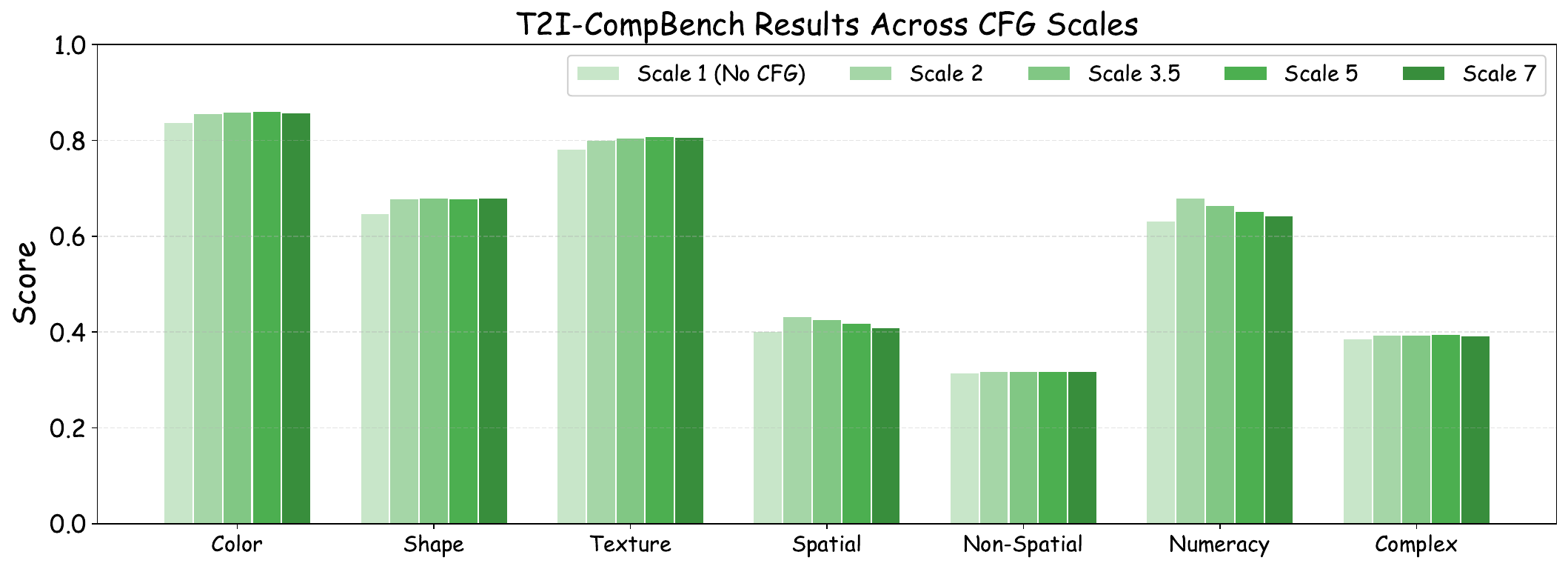}
    \vspace{-0.55cm}
    \caption{Comparison of SDv3.5 performance across different CFG scales after APO fine-tuning. Lower guidance scales (e.g., $2$) yield stronger compositional alignment and overall image quality.}
    \label{fig:cfg_adjust_t2i_compbench_results}
    \vspace{0.4cm}
    \includegraphics[width=\linewidth]{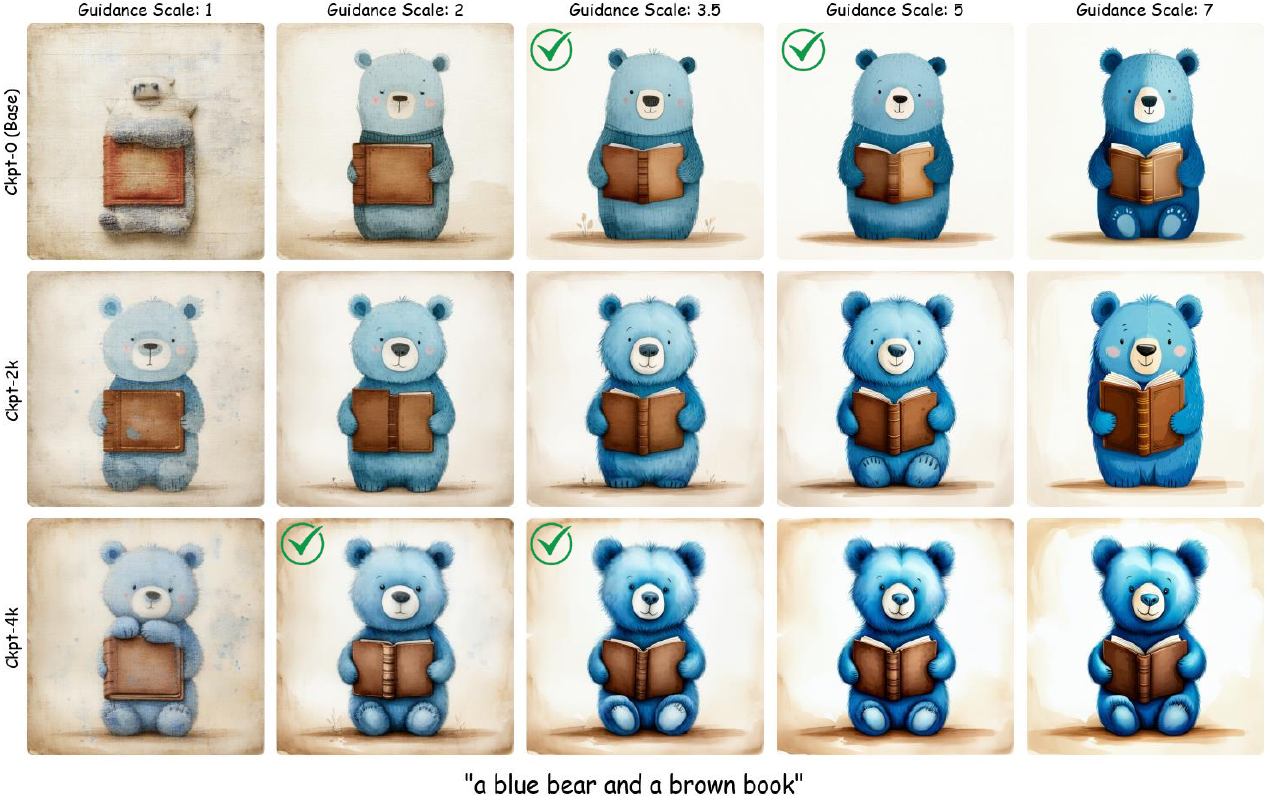}
    \vspace{-0.55cm}
    \caption{Qualitative comparison of SDv3.5 generations at different CFG scales during APO fine-tuning. After fine-tuning, higher guidance scales lead to oversaturated and degraded images, while lower scales preserve fidelity.}
    \label{fig:cfg_adjust_qualitative}
\end{figure}

\begin{figure*}[t]
    \centering
    \includegraphics[width=0.9\linewidth]{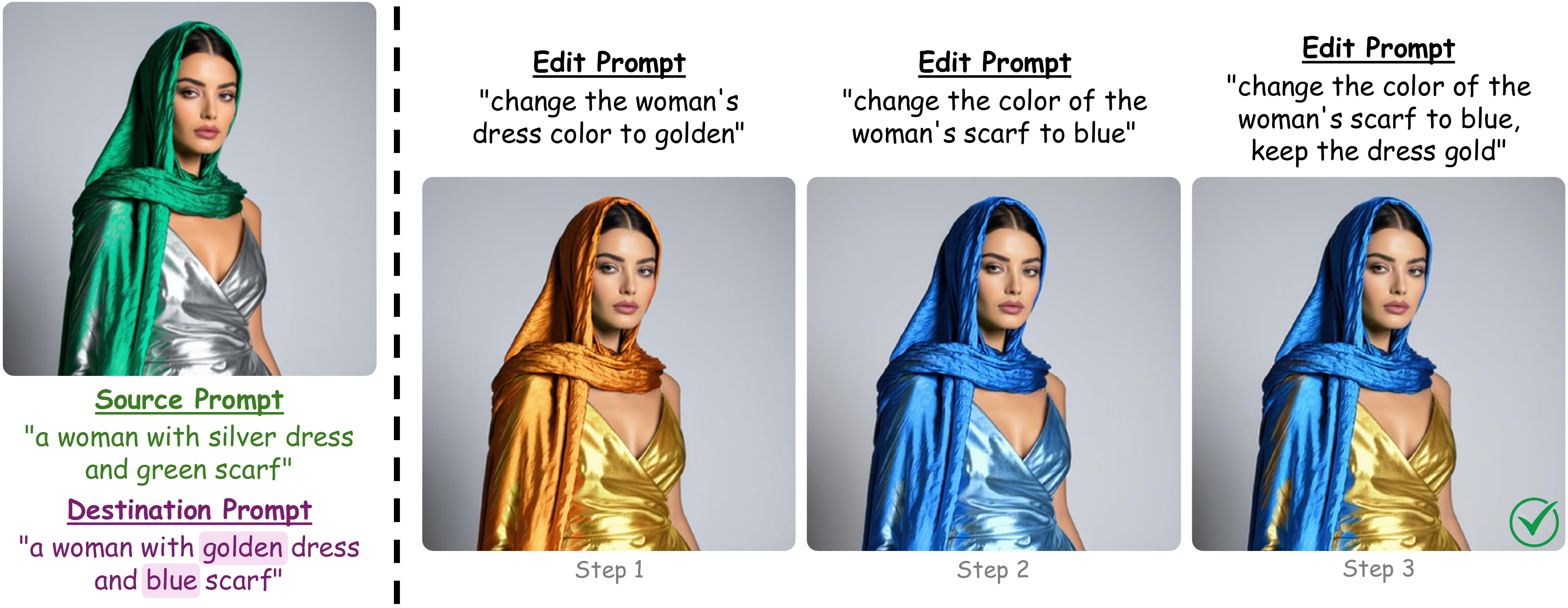}
    \vspace{-0.20cm}
    \caption{\textbf{Image Editing Agent Example Scenario.} Given a source image, its prompt, and a target prompt, the image-editing agent leverages editing tool and VQA to produce a correct contrastive sample. Although the editing tools may introduce unintended modifications (Steps 1 and 2), the agent detects these errors through reasoning and VQA feedback, adjusts its intermediate prompts, and ultimately generates the intended result.}
    \label{fig:image_edit_agent_example}
\end{figure*}

Interestingly, the latter strategy—using a smaller $\beta$ for lower distances and a larger $\beta$ for higher distances—produced the strongest performance. This behavior is reminiscent of “hard-negative’’ emphasis in SVMs, where samples close to the decision boundary carry more informative gradients. In our setting, trajectories with subtle compositional differences from the positive sample represent such hard negatives, and prioritizing them leads to more effective preference optimization.

\subsubsection{Classifier Guidance Scale Adjustment}

During fine-tuning with our APO objective, the unconditional model is no longer being trained. As a result, in models that rely on explicit classifier guidance at inference time (e.g., SDv3.5), the effective classifier-free guidance (CFG) behavior shifts, requiring us to re-adjust the guidance scale post-training. Figure~\ref{fig:cfg_adjust_qualitative} illustrates this effect: using a fixed random seed but varying CFG scales across APO training steps, we observe that the model’s signal becomes stronger at low scales (e.g., CFG = 1). Consequently, higher CFG scales lead to oversaturation and degraded image quality. This suggests that the guidance scale should generally be reduced after APO fine-tuning. Figure~\ref{fig:cfg_adjust_t2i_compbench_results} reports T2I-CompBench performance at iteration 6k for SDv3.5 under different CFG scales. As shown, a scale of 2 achieves the best overall performance among the tested values.

\subsubsection{Agent Ablation}

In this section, we expand on the ablation study of the image-generation agent discussed in Section~\ref{sec:ablations}. To better understand the agent’s behavior across different settings, we examine its tool-use patterns and reasoning steps throughout the intermediate stages of generation. Table~\ref{tab:agent_ablation_stats_appendix} summarizes statistics such as the number of tool calls under different configurations and the number of images produced by the agent.

Several observations emerge from this analysis. Smaller LLMs within the agent struggle to invoke tools correctly and frequently fail to produce valid images. Stronger image-generation models such as SDv3.5 lead to fewer VQA and editing calls, which aligns with the improved baseline generation quality. In addition, using a more capable VQA model yields more reliable image assessments, enabling the agent to identify flaws more accurately, reason more effectively during the generation process, and produce fewer but higher-quality final outputs.

\subsubsection{Dataset scale and Human Filtering}

\begin{table}[t]
\vspace{-0.5cm}
\centering
\small
\caption{Comparison of SDXL performance on agent-generated datasets with and without human filtering.}
\vspace{-0.3cm}
\begin{tabular}{cccc}
\toprule
\textbf{Dataset Scale} & \textbf{Curated} & \textbf{Dataset Size} & \makecell{\textbf{T2I-CompBench}\\\textbf{(Numeracy)}} \\
\midrule
\multirow{2}{*}{Small} & \xmark & $\sim500$ & 0.7221 \\
                       & \cmark & $\sim800$ & 0.7344 \\
\midrule
\multirow{2}{*}{Big}   & \xmark & $\sim2000$ & 0.7386 \\
                       & \cmark & $\sim1100$ & 0.7317 \\
\bottomrule
\end{tabular}
\label{tab:dataset_human_filtering}
\end{table}

In this section, we assess the quality of the agent-generated dataset and examine the impact of manual human filtering. We construct two variants of a numeracy-focused dataset (targeting counting and numerical reasoning from T2I-CompBench): a smaller set of roughly 800 samples and a larger extended set of approximately 2,000 samples. For each, we manually inspect all clusters and retain only high-quality examples, yielding filtered subsets of about 500 and 1,100 samples, respectively.

We fine-tune SDXL under identical training conditions (objective, batch size, and hyperparameters) on all four datasets. The results, shown in Table~\ref{tab:dataset_human_filtering}, reveal an interesting pattern: while manual filtering improves performance in the small-dataset regime, simply scaling up the agent-generated dataset yields similar or better improvements without any human intervention. This highlights the effectiveness and scalability of our agentic data generation pipeline, without any human any the loop.

\newpage

\newtcolorbox{promptbox}{
  colback=gray!5!white,
  colframe=black!50,
  boxrule=0.5pt,
  arc=3pt,
  outer arc=3pt,
  boxsep=4pt,
  left=4pt,
  right=4pt,
  top=4pt,
  bottom=4pt,
  fontupper=\ttfamily\small,
}

\begin{figure*}[t]
\centering
\begin{promptbox}
\textbf{Image Generation Agent Instructions}\\\\
Your task is to generate an image that accurately reflects the given prompt. Because the image generation tool may not always capture every detail correctly, you must verify each result using the visual question answering (VQA) tool. For example, if the prompt is "a cat playing with a green ball and two frisbees", you should ask targeted questions such as "How many cats are there?", "How many balls are there?", "What color is the ball?", "How many frisbees are present?", and "What color are the frisbees?" to confirm that the image matches the description. \\

Start by attempting up to three independent generations. After each generation, use the VQA tool to verify whether the image fully matches the prompt. If one of these generations is correct, return its ID immediately. If none are correct but one is close, select the best among them and refine it using the editing tool. \\

The editing tool allows you to fix errors step by step. For example, if the generated image shows "a cat playing with two green balls" instead of one, you can instruct the editing tool to "remove one of the balls". Similarly, if the prompt is "a dog with a black hat and red sunglasses" but the image only shows "a dog", you can first edit the image to "add a hat to the dog", verify it with VQA, and then make another edit to "add red sunglasses to the cat". If the sunglasses turn out purple instead of red, you can refine them by asking the editing tool to "change the color of the sunglasses to red". Another example is when the prompt is "8 apples". Models often struggle with exact counts, so you may start with an image that contains, for example, six apples and then add one apple at a time, verifying after each edit until the correct count is reached. \\

After every edit, you must run a full verification using VQA. Do not restrict verification to just the edited object. Ask complete questions about all the important entities in the image (e.g., "What color is the hat?", "What color are the sunglasses?") and compare the answers with the ground-truth information you have. This ensures that the intended change was applied correctly and that no other colors, attributes, or objects were accidentally altered. \\

Do not make more than 10 attempts in total, where each attempt may be either a fresh generation or an edit of a previously generated image. At the end of this process, return the ID of the verified image that fully satisfies the prompt. If no image meets the requirements within the allowed attempts, return -1.
\end{promptbox}
\vspace{-0.4cm}
\caption{Instructions for Image Generation Agent}
\label{fig:image_generation_agent_instructions}
\end{figure*}

\begin{figure*}[t]
\centering
\begin{promptbox}
\textbf{Image Editing Agent Instructions}\\\\
You will be provided with three pieces of information at the start: \\
- An image ID in the database (Image IDs can be used with both the image editing tool and the VQA tool). \\
- A source prompt that describes the content of the initial image. \\
- A destination prompt that describes the target image you need to produce. \\

Your task is to determine and describe the sequence of image editing operations required to transform an image that corresponds to a source prompt into one that matches a destination prompt. \\

You have access to an image editing tool. Each operation you design must be simple and atomic, meaning it should represent a small, straightforward change (e.g., changing a color, removing one object, or adding one object). This makes it more likely that the editing tool can perform the change successfully.
After each edit, you must verify whether the change was applied correctly by using the visual question answering (VQA) tool. Also check that no unintended parts of the image were altered. Editing tools sometimes disrupt unrelated elements, so use multiple VQA questions to validate the integrity of the whole scene based on the prompt and edits.   \\

For example, suppose at some point in the editing process you have an image corresponding to "a red book and a yellow vase", and the editing instruction is "make the vase green". After generating the edited output, you should ask verification questions such as "how many books are there?", "what color are the books?", "how many vases are there?", and "what color are the vases?". In general, every object, attribute, and relation described in the prompt—both before and after the edit—should be explicitly tested. No need to ask questions like "Are there any unusual colors or changes to the vase or the scene?", since the vqa model only sees the new image. \\

If the edit is correct and the rest of the image is intact, continue to the next required operation. If the edit fails or introduces unwanted changes, retry, adjust your operation, or attempt a different editing path. \\

Examples (starting from the source prompt: "a red book and two yellow vases"): \\
- Destination: "a green book and two yellow vases" -> (edits: 1) change the book's color to green) \\
- Destination: "a book and two yellow vases" -> (edits: 1) change the book's color to blue) [note: just a color other than red] \\
- Destination: "two yellow vases" -> (edits: 1) remove the book) \\
- Destination: "a red book and a yellow vase and a green vase" -> (edits: 1) change the color of one of the vases to green) \\
- Destination: "a red book and a yellow vase" -> (edits: 1) remove one yellow vase) \\
- Destination: "a red book and two blue vases" -> (edits: 1) change the color of the vases to blue) \\
- Destination: "two purple vases" -> (edits: 1) remove the book, 2) change the color of the vases to purple) \\

Some notes: \\
- Do not make more than 10 calls to the image editing tool. \\

Final Output: \\
If you successfully transform the image into one that matches the destination prompt, print [success].  \\
If you cannot achieve the transformation within the allowed edits, print [failed]. \\

\end{promptbox}
\vspace{-0.4cm}
\caption{Instructions for Image Editing Agent}
\label{fig:image_edit_agent_instructions}
\end{figure*}

\end{document}